\documentclass{article}

\usepackage[preprint]{neurips_2025}

\usepackage[utf8]{inputenc} %
\usepackage[T1]{fontenc}    %
\usepackage{hyperref}       %
\hypersetup{
  colorlinks,
  linkcolor={red},
  citecolor={blue},
  urlcolor={magenta}
}
\usepackage{url}            %
\usepackage{booktabs}       %
\usepackage{amsfonts}       %
\usepackage{nicefrac}       %
\usepackage{microtype}      %
\usepackage{xcolor}         %

\usepackage{graphicx}       
\usepackage{amsmath, amsfonts, amssymb}
\usepackage{algorithm}
\usepackage{algpseudocode}
\usepackage{amsthm} 
\usepackage{amsmath}
\usepackage{thmtools,thm-restate}
\usepackage{enumitem}
\usepackage{multirow}
\usepackage{makecell}
\usepackage{booktabs}
\usepackage{wrapfig}
\usepackage{subcaption} 
\usepackage{lipsum}
\usepackage{tabularx} 
\usepackage{array} 

\newcommand{\modelname}{\textit{SparseSSM}}
\newtheorem{thm}{Theorem}
\newtheorem{lemma}{Lemma} 
\allowdisplaybreaks[4]

\title{\modelname: Efficient Selective Structured State Space Models Can Be Pruned in One-Shot}

\author{%
  Kaiwen Tuo \\
  Tongji University\\
  Shanghai, China 200092 \\
  \texttt{cfintuo@gmail.com} \\
  \And
  Huan Wang \\
  Westlake University \\
  Hangzhou, China \\
  \texttt{wanghuan@westlake.edu.cn} \\
}

\begin{document}
\maketitle

\begin{abstract} \label{sec:abstract}

  State-space language models such as Mamba match Transformer quality while permitting linear complexity inference, yet still comprise billions of parameters that hinder deployment. 
  Existing one-shot pruning methods are tailored to attention blocks and fail to account for the time-shared and discretized state-transition matrix at the heart of the selective state-space module (SSM). 
  In this paper, we introduce \modelname, the first training-free pruning framework that extends the classic optimal brain surgeon (OBS) framework to state space architectures.
  Our layer-wise algorithm \textbf{(i)} derives an approximate second-order saliency score that aggregates Hessian-trace information across time steps, \textbf{(ii)} incorporates a component sensitivity analysis to guide feed-forward network (FFN) pruning, which also sheds light on where redundancy resides in mamba architecture, \textbf{(iii)} can be easily extended to semi-structured and structured sparsity.
  Empirically, we prune 50\% of SSM weights without fine-tuning and observe no zero-shot accuracy loss, achieving the current state-of-the-art pruning algorithm for Mamba-based LLMs.
  
\end{abstract}

\section{Introduction} \label{sec:intro}

The rapid expansion of Transformer-based large language models (LLMs), which now scale to hundreds of billions of parameters \citep{LLaMA, opt, bloom}, has created an urgent demand for efficient model compression.
The deployment of such models involves substantial computational cost and environmental impact.
Among various compression techniques, network pruning, the removal of redundant weights, remains a classic yet effective method to reduce model size and accelerate inference with minimal performance degradation \citep{obd, obs, deep-compression, llm-pruner, sparsegpt}.
However, many pruning approaches, especially those based on magnitude or gradient information, require retraining or fine-tuning to recover accuracy \citep{magnitude, pruning-filters, channel-pruning}, which is feasible for smaller models but becomes prohibitively expensive on the scale of modern LLMs.
To address this, researchers have introduced \textbf{training-free} pruning strategies that induce sparsity in one shot without any additional optimization, in addition to more traditional \textbf{training-based} pipelines when budget allows \citep{snip, synflow, obc}.

In the Transformer regime, lots of methods have emerged to support one-shot pruning without fine-tuning, achieving surprisingly strong performance. 
Notably, SparseGPT \cite{sparsegpt} introduced an approximate optimal brain surgeon (OBS) \citep{obs} inspired framework that prunes massive LLMs to over 50\% sparsity in a single pass with negligible degradation. 
SparseGPT leverages local second-order information to reconstruct pruned weights and minimize output error. 
Simpler alternatives such as Wanda \citep{wanda} offer lightweight heuristics based on the product of magnitude and input activation per output neuron, yet still achieve accuracy competitive with more sophisticated OBS-based approaches, all without retraining.
These methods exemplify a growing trend in Transformer-based LLMs compression: \textbf{OBS-guided pruning} that can operate efficiently at scale while preserving LLM quality, even at high sparsity levels \citep{ALPS}. 

Recently, Mamba \citep{mamba, mamba2} has emerged as a promising state space alternative to Transformers, replacing the attention mechanism with a \textbf{selective state space model (SSM)} which enables linear complexity sequence processing and significantly faster inference.
Mamba-based LLMs have demonstrated competitive performance compared to Transformer-based LLMs of similar scale \citep{falcon-mamba}.
Despite their efficiency and effectiveness, Mamba-based LLMs still contain billions of parameters and thus stand to benefit greatly from pruning. 
To date, most pruning research has focused on Transformer-based models, with limited efforts targeting Mamba or other state-space architectures.
This motivates a key question: \textbf{Can we design a training-free pruning method specifically tailored to the SSM module in Mamba?}

However, most existing pruning techniques developed for feed-forward and attention layers of Transformers cannot be directly transferred to the SSM module of Mamba, due to its time-shared parameters and discretization operation.
For example, \textbf{(i)}  The parameter $ A $ in the SSM module is time‑shared, meaning that any pruning decision must account for importance metrics computed at each time step; unlike spatial aggregation, however, the activations at one step are directly influenced by the previous time step.
\textbf{(ii)} During execution, $ A $ is discretized into $ \Delta A $; therefore, pruning must explicitly consider this discretization operation. 

\begin{figure}[t]  
  \centering   
  \includegraphics[
    trim=2.4cm 3cm 2.6cm 3cm,
    clip,
    width=1.0\linewidth
  ]{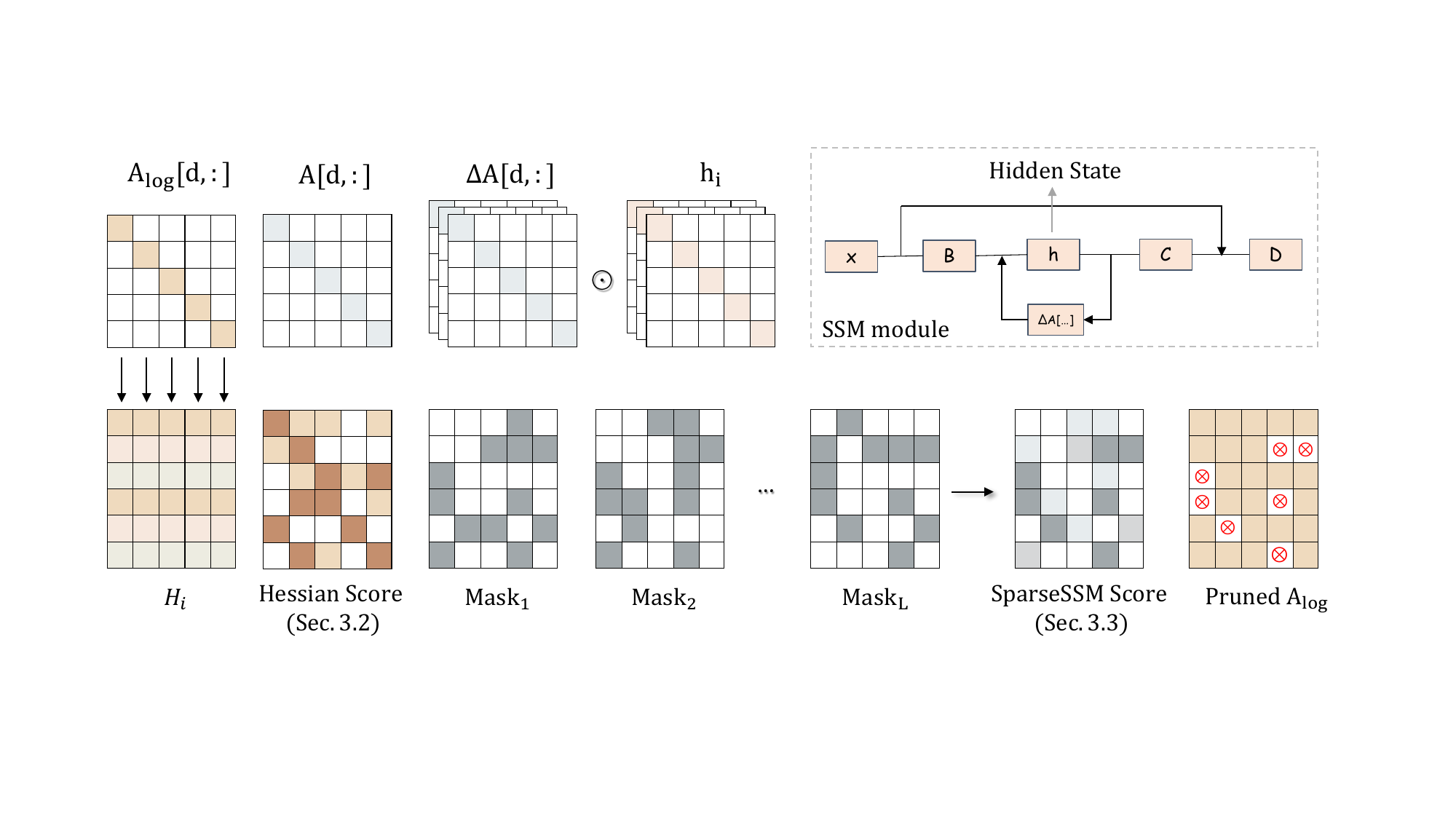}
  \caption{Illustration of \modelname. The \textbf{first row} depicts the evolution of the diagonal parameter matrix $A_{log}$ within the SSM module in Mamba, together with a schematic of the forward-propagation process. In the \textbf{second row}, the \textbf{left panel} shows the procedure for obtaining a mask from the Hessian estimate at a single time step (see Section~\ref{sec:method_hessian}), while the \textbf{right panel} presents our strategy for merging the masks across all time steps (see Section~\ref{sec:method_importance}).}
  \label{fig:overview}
\end{figure}

To address these issues, we proposed \modelname, a layer-wise pruning method that generalizes the traditional OBS framework to efficient selective structured state space models (see Fig.~\ref{fig:overview}). 
Our technical \textbf{contribution} can be summarized as follow:

\begin{itemize}[noitemsep,topsep=0pt,parsep=0pt,partopsep=0pt, leftmargin=*]
    \item We introduce \modelname\ that adapts the classic optimal brain surgeon framework to the selective SSM module in Mamba. Our method computes approximate second-order weight importance for the time-sharing SSM parameters, enabling principled one-shot pruning of the SSM layers. This is the first application of OBS-based pruning to Mamba’s architecture, addressing the challenges of its discrete diagonalized design.
    \item  We further improve \modelname\ with two complementary techniques. First, we propose a mask aggregation method to address the time-sharing nature of the SSM module. Second, we provide an in-depth analysis of the components of Mamba and compare their pruning tolerance, which informs the FFN pruning strategy, guiding which linear projections should be pruned more conservatively, and sheds light on where redundancy resides in Mamba.
    \item \modelname\ achieves significantly superior performance compared to current state‑of‑the‑art pruning algorithms for Mamba-based LLMs. Through experiments on standard language modeling benchmarks, we demonstrate that our method can prune 50\% of the SSM weights without performance degradation for Mamba-370M, also without fine-tuning or calibration. We also adapt \modelname\ to semi-structured and structured sparsity format.
\end{itemize}

\section{Related Work} \label{sec:related_work}

\textbf{Selective State Space Models.\,\,\,\,} Selective State Space Models (SSMs) have emerged as promising alternatives to the attention layers in Transformers, particularly due to their computational complexity in linear time and their ability to handle long-range dependencies efficiently \citep{mamba, mamba2}.
Unlike traditional attention-based mechanisms \citep{transformer}, whose complexity grows as the square of the sequence length, SSMs operate linearly, allowing efficient processing of exceptionally long sequences \citep{hippo, s4, s5}.
Mamba allows parameters within SSM layers to dynamically vary based on the input sequence \citep{mamba}, while Mamba-2 further introduces State Space Duality (SSD) to improve computational parallelism and hardware utilization \citep{mamba2}. 
Building upon these foundations, Falcon Mamba presents a new foundational LLM based on the Mamba architecture \citep{falcon-mamba}.
Recent hybrid architectures combining SSMs with Transformers have further demonstrated significant empirical gains, exploiting the complementary strengths of both architectures \citep{jamba, zamba, simba, hymba}.

\textbf{Network Pruning in LLMs.\,\,\,\,}
Network pruning is a widely adopted technique to reduce the computational cost and memory footprint of deep neural networks by eliminating redundant parameters \citep{obd, deep-compression}.  
Applying pruning techniques to Large Language Models (LLMs) presents unique challenges compared to smaller models like convolutional networks \cite{woodfisher, chita} or even moderate-sized language models like BERT \cite{bert}.
To address these challenges, several efficient pruning methods for LLMs have been proposed.
Regarding granularity, these pruning methods for LLMs can be either unstructured, targeting individual weights \citep{sparsegpt, wanda, llm-surgeon}, or structured, removing entire units like channels, filters, or attention heads \citep{llm-pruner, darwinlm}.
Our work mainly focuses on one-shot unstructured pruning due to its efficiency and potential for high sparsity, while it can also be extended to structured patterns.

\textbf{Layer-wise Unstructured Pruning Methods.\,\,\,\,}
To date, layer-wise pruning methods for LLMs are primally based on the optimal brain surgeon (OBS) \citep{obs} framework.
OBC \citep{obc} proposed the ExactOBS algorithm to reduce computational burden, reformulating layer-wise pruning as a row-wise operation. 
To address the massive parameters of LLMs, SparseGPT \citep{sparsegpt} further tackles the expensive Hessian computation by employing partial weight updates and adaptive mask selection.
Other techniques have explored more aggressive Hessian estimation \citep{wanda}, extension to structured sparsity \citep{slimgpt, structured-obs}, and methodological improvement for better performance \citep{i-obs, c-obs, ALPS}.

\textbf{Pruning Methods for Mamba.\,\,\,\,} While pruning algorithms tailored for Transformer-based LLMs have achieved considerable success, pruning Mamba architectures \citep{mamba, mamba2} still encounter substantial challenges.
Gwak et al. reveal the redundancy and compressibility of state space models, thereby motivating the application of pruning techniques to SSM architectures \citep{kwak2024layer}.
Some early studies have focused on structured pruning of Mamba, such as the coarse‐grained removal of SSM modules or whole blocks by Mamba-Shedder \citep{mamba-shedder} —and on unstructured pruning, evaluating a variety of pruning techniques applied to the Mamba architecture \citep{mambap}.
Nearly, Taghibakhshi et al. propose a group-aware pruning strategy tailored for hybrid attention-SSM models, which simultaneously combines SSM components alongside other network elements such as FFN neurons \citep{hybridp}.
Compared to earlier strategies, our solution departs in two critical aspects: \textbf{(i)} \modelname\ extends the classic OBS framework to address pruning in the SSM module, providing rigorous theoretical justification and comprehensive experimental validation.
\textbf{(ii)} We propose a one‑shot, unstructured pruning algorithm for Mamba that requires no fine‑tuning.

\section{Method} \label{sec:method}

In this section, we demonstrate how the OBS framework can be adapted to the Mamba architecture and present our method, \modelname.
To start, we provide a detailed overview of Mamba’s forward‐propagation pipeline, emphasizing the internal computations within its SSM modules.
We then describe our targeted Hessian‐matrix calculation technique and derive the resulting importance metrics. 
Finally, we explore pruning strategies for the feed‐forward network (FFN) layers.

\subsection{Forward Propagation Pipeline of Mamba} \label{sec:method_forward}

We first dive deeper into the forward propagation of a single mamba block.
The Mamba architecture decomposes into two complementary components: a feed‑forward network (FFN) that performs feature projection and preliminary transformation, and a state space model component that selectively captures and processes sequential dependencies.

State space models (SSMs) provide a sequence modeling paradigm based on latent state dynamics. 
In Mamba's SSM layer, the state $\mathbf{h}_t \in \mathbb{R}^{B \times D \times N}$ evolves recurrently with input $\mathbf{x}$ as: 
\begin{equation}
  \mathbf{h}_t \colon=\; \widehat{A} \mathbf{h}_{t-1} + \widehat{B}\,\mathbf{x}_t,
  \quad
  \mathbf{y}_t \colon=\; C^\top \mathbf{h}_t ,
  \label{eq:ssm}
\end{equation}
where $\widehat{A} $ denotes the state transition matrix , $\widehat{B}$ and $C$ are parameters of SSM, and $\mathbf{y}_t$ is the output. 
Mamba achieves efficient sequence processing by making $A_t$ diagonal, while achieving selectivity by instantiating a dedicated SSM for each token. 
Specifically, the original input and output gate matrices \(B\) and \(C\) are expanded to shape \(\mathbb{R}^{B\times L\times N}\), and the transition parameter \(A\) is held via zero‑order‑hold and discretized into $\Delta A \;\in\;\mathbb{R}^{B\times L\times D\times N}\ $, thereby endowing each batch ($B$), each sequence position ($L$), and each channel ($D$) with its own independent SSM instance. The discretization and parameterization procedures for $A$ are, respectively, as follows:
\begin{equation}
    (\Delta A)_{b,\ell,d,n} \;=\; \exp\ \!\bigl(\delta_{b,\ell,d}\,A_{d,n}\bigr),
    \qquad
    (A_{log})_{d,n} \;=\; -\log\ \!\bigl(A_{d,n}\bigr),
    \label{eq:discrete}
\end{equation}
where $ \delta_{b,\ell,d} $ denotes element of the stride $ \Delta \in \mathbb{R}^{B \times L \times D} $.

This selective, input-dependent design addresses the limitations of earlier linear time-invariant SSMs and enables long-context reasoning by dynamically controlling which state dimensions carry information.
However, its recurrent structure and discretization operations render prior methods inapplicable to pruning the parameters of the SSM, specifically the $A$ matrix.

In the Mamba layer, we leverage the selective scan algorithm to traverse the sequence and record the internal state contributions. 
At each token $t$, we denote by $ \mathbf{h}_{t} \in \mathbb{R}^{B \times D \times N} $ the hidden state tensor at time step $t$, containing the activation values for every batch and channel.
This internal signal $\mathbf{h}_t$ reflects how much each state dimension $i$ remembers its past activation at step $t$. 
By collecting these values across all time steps $t=1\dots L$ for a given layer, we obtain token-wise activation statistics for each state dimension. 
In this way, the selective scan provides a direct window into the layer’s memory utilization, which we will exploit to guide pruning.

\subsection{Hessian Matrix Estimation of SSM Layer} \label{sec:method_hessian}

To formally quantify each parameter’s importance in SSM layers, we adopt a Hessian-based analysis inspired by optimal brain surgeon (OBS) pruning \citep{obs, sparsegpt}.
The goal of pruning is to identify a sparse weight matrix $A_{log}$ that minimizes the reconstruction error between the original and pruned layer outputs. 
Let $\mathbf{SSM}$ denote Mamba's SSM layer, then the problem can be formulated as:
\begin{equation}
  \arg\min_{A'}
  \big \lVert
  \mathbf{SSM} \bigl(A, \theta, x \bigr)
  - 
  \mathbf{SSM} \bigl( A^\prime, \theta, x \bigr)
  \big \rVert_2^2 ,
  \label{eq:target}
\end{equation}
where $\theta$ represents the output of the formal linear projection. 
Based on OBS framework, pruning of parameter $A$ requires no compensation because it is essentially a concatenation of multiple diagonal matrices, whose elements' importance can be defined as $ \varepsilon_{m} = w_{m}^{2} H_{mm}\, $, where $H_{mm}$ denotes the $m$-th diagonal element of the Hessian matrix.
It actually measures the curvature of the loss concerning the parameter $m$.
However, precisely computing the entries of the Hessian matrix for an SSM module is challenging, because for a given input x, the SSM may be unrolled across time steps as follows:
\begin{equation}
  \begin{aligned}
  h_i &= \Delta A_i \odot h_{i-1} + \Delta B_i \odot x\,, 
  \quad h_{-1} = 0\,,\\
  y_i &= h_i^{\!\top} C_i 
       = \sum_{j=0}^{i}
         \Bigl[\bigl(\prod_{k=j+1}^{i}\Delta A_k\bigr)
                \odot \Delta B_j \odot x_j\Bigr]^{\!\top} C_i\,,
  \end{aligned}
  \label{eq:unroll}
\end{equation}
where $x$ denotes the input of the SSM module, and $y_i$ denotes the output at each time step $i$.
To address this problem shown in Eq.~(\ref{eq:unroll}), we consider computing the Hessian matrix at each time step and using the hidden state to assess the importance of elements.
For the SSM module, we consider the loss $\mathcal{L}$ incurred when it processes input data,  which operates in a way analogous to Backpropagation Through Time (BPTT) \citep{bptt}.
\begin{equation}
    \mathcal{L} = \frac{1}{B} \sum_{b=1}^{B} \ell(\mathbf{y}_{b,0:L-1})
     = \frac{1}{B} \sum_{b=1}^{B} \lVert \mathbf{y}_{b,0:L-1}-\mathbf{\hat{y}}_{b,0:L-1} \rVert_2^2,
    \label{eq:loss}
\end{equation}
where $L$ denotes the length of the input sequence and also the total time step.
Based on this hypothesis, we propose Theorem~\ref{thm:ssm-obs} to give an estimation of the Hessian matrix, which, to the best of our knowledge, represents a novel theoretical perspective for analyzing and guiding pruning.
\begin{thm}[SSM Hessian Matrix Estimation]\label{thm:ssm-obs}
Let \(\displaystyle A_{\log}\in\mathbb R^{D\times N}\)
be the matrix of parameters for SSM diagonals, \(\{\delta_{b,i,d}\}\) the discretization increments, and \(\{h_{b,i-1,d,n}\}\) the hidden activations before step \(i\).  
Under the diagonal character of parameter $A_{log}$, the per‑parameter importance score is:  
\begin{equation}
I^{\log}_{d,n}
\;=\; A_{\log,d,n}^{2}\,\kappa \sum_{b,i} h_{b,i-1,d,n}^{2}\,
\delta_{b,i,d}^{2}\, A_{d,n}^2 e^{2\,\delta_{b,i,d}\,A_{d,n}}
\;\propto\;
A_{\log,d,n}^{2}
\;\sum_{b,i}h_{b,i-1,d,n}^{2}\,.
\end{equation}
In other words, after absorbing \(\kappa\) and the mean effect of \(\delta_{b,i,d}^{2}e^{2\delta_{b,i,d}A_{d,n}}\) into a global constant, the OBS ranking of each elements reduces to the simple product
\(\;A_{\log,d,n}^{2}\,\times\sum_{b,i}h_{b,i-1,d,n}^{2}\).
\end{thm}

\subsection{Importance Estimation for Integrated Time Steps} \label{sec:method_importance}

Our proposed Theorem~\ref{thm:ssm-obs} enables the Hessian matrix of the SSM module to be quickly and accurately estimated. 
However, pruning $A_{log}$ remains challenging due to its parameter time-sharing property. 
This implies that while the activation at each time step can produce a pruning mask for $A_{log}$, pruning at one time step affects the selection of the pruning mask at the subsequent time step. 
Consequently, merging these masks becomes problematic.

To merge these pruning masks produced by each time step, we propose a hierarchical aggregation protocol.
This protocol, detailed in Algorithm~\ref{alg:ts_prune}, goes beyond simple mask merging by using a deferred commitment approach, where saliency evaluations at each time step are combined to establish a globally consistent pruning criterion.
The initial stage of our protocol involves the identification of potential pruning candidates at each time step, then we employ the pruning frequency as the definitive pruning criterion.
Algorithm~\ref{alg:ts_prune} provides the detailed steps of our proposed \modelname\ method.
\begin{algorithm}[H]
  \caption{Time-Selective One-Shot OBS Pruning for the SSM Matrix $A_{\text{log}}$}
  \label{alg:ts_prune}
  \begin{algorithmic}[1]

\Statex{\bfseries Phase 1: Statistic accumulation}
\State $\mathbf{S}\gets\mathbf{0}_{L\times D\times N}$,\; $n\gets0$
\For{each mini-batch $b=1,\dots,B$}
    \State Run the forward pass of the layer and collect
    $\mathbf{h}^{(b)}_{\text{1:L}}$ 
    \State $n\gets n+1$
    \State $\mathbf{S}\gets\dfrac{n-1}{n}\,\mathbf{S}+ 
            \dfrac{1}{n}\sum_{t=1}^{L}\!\left(\mathbf{h}^{(b)}_{t}\right)^2$
            \label{line:stat_update}
\EndFor
\Statex{\bfseries Phase 2: Per-time-step candidate selection}
\State $K\gets\bigl\lceil pDN\bigr\rceil, \,\mathbf{C}\gets\mathbf{0}_{D\times N}$ 
\For{$t=1,\dots,L$} \label{line:time_loop_start}
    \State $\mathbf{M}_t \gets A_{\text{log}}^{\!2}\odot\mathbf{S}_t$
            \Comment{OBS importance score at step $t$}
    \State $\mathcal{I}_t \gets
            \operatorname{arg\,smallest}_K\!\bigl(\mathbf{M}_t\bigr)$
            \label{line:topk}
    \State $\mathbf{C}_{\mathcal{I}_t}\gets\mathbf{C}_{\mathcal{I}_t}+1$
\EndFor \label{line:time_loop_end}

\Statex{\bfseries Phase 3: Final mask construction}
\State $\mathcal{I}_\star\gets
        \operatorname{arg\,largest}_K\!\bigl(\mathbf{C}\bigr)$
        \Comment{most frequently selected indices}
\State $\mathbf{M}\gets\mathbf{1}_{D\times N}$;\;
        $\mathbf{M}_{\mathcal{I}_\star}\gets 0$ \label{line:mask}
\State $\widetilde{A}_{\text{log}}\gets A_{\text{log}}\odot\mathbf{M}$ 
\end{algorithmic}
\end{algorithm}

\subsection{Sensitivity-Aware Pruning of the FFN Component} \label{sec:method_ffn}

\begin{wrapfigure}{r}{0.5\textwidth}
  \centering
  \includegraphics[trim=0.4cm 0.1cm 0.4cm 0.1cm,
    clip, width=0.46\textwidth]{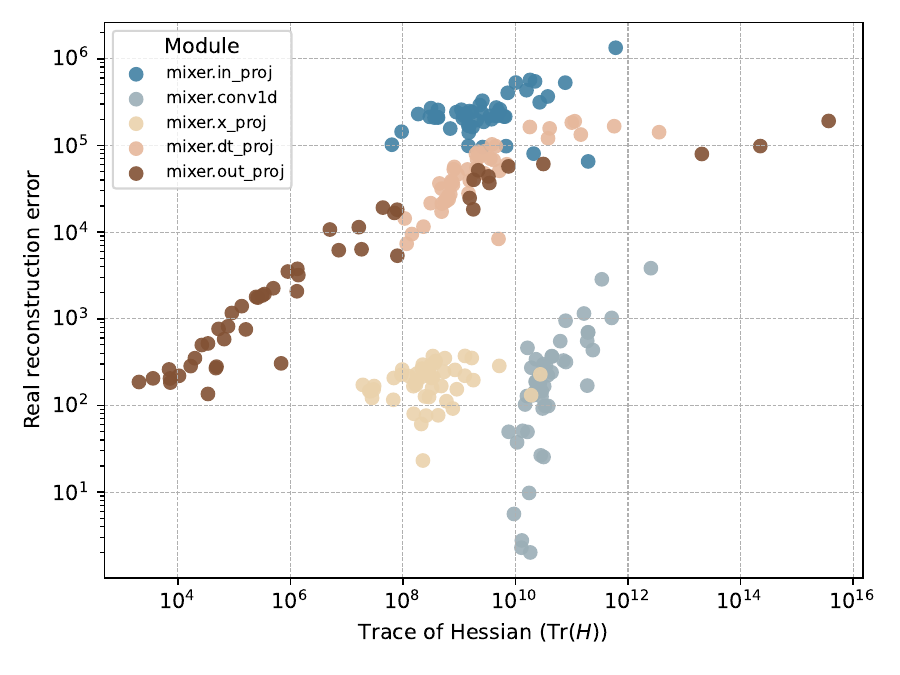}
  \caption{ The Hessian matrices and corresponding reconstruction errors for each module of the Mamba-370M FFN at a sparsity level of 50\%. Different modules are represented by different colors.}
  \label{fig:ffn_sensitivity}
\end{wrapfigure}

While our primary contribution is the OBS-based pruning of Mamba’s state space module, we also perform pruning on the standard feed-forward networks (FFNs) in the model to further reduce parameters based on the SparseGPT \citep{sparsegpt} framework. 
The forward‐propagation blocks in Mamba are composed primarily of linear layers and one‐dimensional convolutional layers. 
Inspired by \citep{sensitive}, we conducted a module‐wise pruning analysis within the feed‐forward network (FFN) and found that their pruneability varies substantially (see Appendix~\ref{sec:details_different_modules}).
In particular, pruning the \texttt{in\_proj} and \texttt{out\_proj} modules incurs a pronounced degradation in overall model performance. 
Moreover, we empirically observe that the reconstruction error of each module grows as its Hessian trace increases, with the rate of this growth varying across modules, as shown in Fig.~\ref{fig:ffn_sensitivity}.

Motivated by these findings, we adopt the sensitivity‐aware pruning framework, treating the \texttt{in\_proj} and \texttt{out\_proj} modules independently. We use the trace of the Hessian matrix of the weights as the sensitivity score, and define the sparsity ratio as:
\begin{equation}
    sparsity = 1 - p - \alpha + \frac{2\alpha\,id}{N - 1},
    \label{eq:sparsity}
\end{equation}
where \(p\) is the target global sparsity, \(\alpha\) defines the allowable deviation interval \(\bigl[1-p-\alpha,\;1-p+\alpha\bigr]\), \(N\) is the total number of weights, and \(id\) is the sensitivity‑rank index of the given weight after sorting by Hessian‑trace importance. 
This formulation ensures that higher‑sensitivity weights (larger \(id\)) are assigned lower sparsity, while exactly satisfying the overall sparsity budget \(p\).

\section{Experiments} \label{sec:experiments}

In this section, we benchmark \modelname\ against leading pruning algorithms for SSM-based LLMs.
The complete experimental protocol and reproducibility details appear in Appendix~\ref{sec:details_setup}. 

\noindent\textbf{Models and Datasets.} \label{sec:ex_models}
We evaluate \modelname\ on the public Mamba checkpoints ranging from 130 million to 1.4 billion parameters \citep{mamba}. 
For all models, we follow the standard calibration protocol on WikiText-2: we randomly sample 128 contiguous segments of 2048 tokens each from the first data shard, as in \citep{sparsegpt}. 
Perplexity is computed as the exponential of the negative log‐likelihood per token, consistent with \citep{ppl}. 
Downstream evaluation uses the raw WikiText-2 validation set \citep{wikitext}, the Penn Treebank corpus \citep{ptb}, and a 10000‐document slice of the C4 validation split \citep{c4}. 
Zero‐shot generalization is measured on PIQA \citep{piqa}, OpenBookQA \citep{openbookqa}, Winogrande \citep{winogrande}, ARC-Easy, and ARC-Challenge \citep{ai2} without any task‐specific fine‐tuning. 
This suite covers both language modeling and reasoning benchmarks, allowing a comprehensive assessment of model performance and generalization.
During implementation, we also referred to mamba-minimal \citep{mamba_minimal} for guidance.

\noindent\textbf{Baselines.} \label{sec:ex_baselines}
We compare \modelname\ against three representative pruning methods under identical calibration and sparsity budgets.
First, global magnitude pruning follows the classical heuristic of removing the smallest‐magnitude weights \citep{magnitude}. 
Second, SparseGPT applies a Hessian‐aware one‐shot pruning strategy \citep{sparsegpt}. However, SparseGPT is not inherently suited to the structural characteristics of SSM modules. Here, we present the results of its naive application.
Third, Mamba-Shedder is a recent selective state space variant tailored for Mamba architectures \citep{mamba-shedder}.
\textit{All baselines and our method share the same configuration to ensure fairness.}

\subsection{Results of Pruning SSM Modules} \label{sec:ex_ssm}
We first isolate the SSM blocks and prune only the learnable diagonal $A_{log}$ matrices. 
Within the state‐space module (SSM), Mamba reparameterizes $A$ via its negative logarithm to enforce $A<0$, thus preserving the module’s robustness. 
Indeed, the parameter $A_{log}$ plays a role analogous to the forget gate in LSTM \citep{lstm} networks and has a profound impact on the predictive capacity of the language model.
 
Table~\ref{tab:ssm-0.5} reports detailed token‐level perplexity and zero‐shot accuracies at 50\% sparsity. 
As demonstrated, our pruning strategy shows excellent efficacy in maintaining SSM stability, even under aggressive sparsification, the module remains well‐conditioned.
For instance, SparseSSM achieves no degradation on most zero-shot tasks and improves zero-shot accuracy by 5.4\% compared to other methods on Mamba-370M. 
The gains stem from our second‐order importance metric combined with a time‐selective mask, as detailed in Section~\ref{sec:method}.
We further observe that pruning the critical parameter $A$ inevitably degrades the generative performance of the model, manifesting as a lower tolerance in perplexity compared to its robustness under zero‐shot evaluation. More results are in Appendix~\ref{sec:details_more_ssm}.

\begin{table}[!htbp]
\centering
\caption{Performance analysis for one-shot unstructured pruning of SSM modules in Mamba models (130M $\sim$ 1.4B) at $50\%$ sparsity. 
Here, $\downarrow$ lower metrics reflect better outcomes, and $\uparrow$ denotes higher metrics reflect better outcomes.
}
\vspace{1mm}
\resizebox{0.95\columnwidth}{!}{%
\renewcommand{\arraystretch}{1.3}
\begin{tabular}{c|c|*{9}{W{c}{3em}}}
\Xhline{3\arrayrulewidth}
Model & Method  & Wiki. $\downarrow$ & PTB $\downarrow$ & C4 $\downarrow$ 
      & OBQA $\uparrow$ & PIQA $\uparrow$ & ARC-e $\uparrow$ & ARC-c $\uparrow$
      & WinoG $\uparrow$ & Avg. $\uparrow$ \\ \hline
\multirow{5}{*}{Mamba-130M} 
  & Dense          & 20.60 & 32.75 & 25.66 & 28.60 & 63.28 & 48.02 & 24.40 & 52.5 & 43.36 \\ 
  & MP \citep{magnitude}            & 740.3 & \underline{1109}  & \underline{273.0} & 26.80 & \underline{58.05} & \underline{39.69} & 22.35 & \textbf{52.33} & \underline{39.84} \\ 
  & Mamba-Shedder \citep{mamba-shedder}  & \underline{698.7} & 1544  & 532.6 & \underline{28.00} & 54.73 & 30.00 & 23.72 & 49.88 & 37.27 \\
  & SparseGPT \citep{sparsegpt}     & 2.4e7 & 6.1e6 & 3.9e5 & 27.60 & 55.28 & 30.64 & \underline{23.98} & 49.25 & 37.35 \\
  & \modelname     & \textbf{27.70} & \textbf{47.81} & \textbf{31.47} 
                   & \textbf{29.20} & \textbf{61.97} & \textbf{44.57} & \textbf{24.40}
                   & \underline{51.60} & \textbf{42.35} \\ \hline 
\multirow{5}{*}{Mamba-370M}
  & Dense          & 14.32 & 23.46 & 19.37 & 31.00 & 68.34 & 54.97 & 27.90 & 55.25 & 47.49 \\ 
  & MP \citep{magnitude}             & \underline{291.2} & 535.9 & \underline{105.0} & \underline{30.40} & 61.70 & 44.23 & 22.61 & 51.38 & 42.06 \\ 
  & Mamba-Shedder \citep{mamba-shedder}   & 334.5 & \underline{446.6} & 221.81 & 23.80 & 54.19 & 29.42 & 23.12 & 52.25 & 36.56 \\
  & SparseGPT \citep{sparsegpt}     & 2696 & 7570 & 613.2 & \underline{30.40} & \underline{65.23} & \underline{49.16} & \underline{25.60} & \underline{52.41} & \underline{44.56} \\
  & \modelname     & \textbf{19.27} & \textbf{31.05} & \textbf{24.72} 
                   & \textbf{32.80} & \textbf{69.64} & \textbf{54.21} & \textbf{27.05} & \textbf{53.67} & \textbf{47.47} \\ \hline 
\multirow{5}{*}{Mamba-790M} 
  & Dense          & 11.96 & 18.45 & 16.62 & 33.80 & 72.63 & 61.07 & 29.44 & 56.27 & 50.64 \\ 
  & MP \citep{magnitude}            & 179.0 & 377.0 & \underline{79.43} & 30.20 & 64.74 & 47.81 & 25.09 & 54.14 & 44.40 \\ 
  & Mamba-Shedder \citep{mamba-shedder}   & 225.48 & 256.32 & 195.47 & 28.20 & 56.47 & 33.29 & 21.50 & 51.07 & 38.11 \\
  & SparseGPT \citep{sparsegpt}     & \underline{110.5} & \underline{242.19}  & 81.87 & \underline{32.80} & \underline{68.34} & \underline{54.42} & \underline{27.47} & \underline{54.93} & \underline{47.59} \\
  & \modelname     & \textbf{14.87} & \textbf{23.81} & \textbf{19.74} 
                   & \textbf{33.40} & \textbf{71.11} & \textbf{58.38} & \textbf{28.16} & \textbf{56.51} & \textbf{49.51} \\ \hline 
\multirow{5}{*}{Mamba-1.4B} 
  & Dense          & 10.75 & 17.05 & 15.17 & 36.40 & 73.88 & 65.57 & 32.85 & 61.17 & 53.98 \\
  & MP \citep{magnitude}             & 100.7 & 190.8 & 54.49 & 30.60 & 67.95 & 53.28 & 24.06 & 52.49 & 45.68 \\ 
  & Mamba-Shedder \citep{mamba-shedder}  & 223.1 & 293.7 & 190.5 & 27.20 & 56.86 & 34.09 & 23.04 & 51.46 & 38.53 \\
  & SparseGPT \citep{sparsegpt}     & \underline{49.77} & \underline{88.20} & \underline{40.74} & \underline{34.40} & \underline{71.38} & \underline{60.10} & \underline{30.03} & \underline{54.78} & \underline{50.14} \\
  & \modelname     & \textbf{14.68} & \textbf{37.79} & \textbf{18.83} 
                   & \textbf{34.80} & \textbf{71.65} & \textbf{62.96} & \textbf{30.97} & \textbf{57.30} & \textbf{51.54} \\
\Xhline{3\arrayrulewidth}
\end{tabular}%
}
\label{tab:ssm-0.5}
\end{table}

\subsection{Results of Pruning the Whole Mamba Architecture} \label{sec:ex_all}

\begin{wrapfigure}{r}{0.5\textwidth} 
  \centering
  \includegraphics[width=0.5\textwidth]{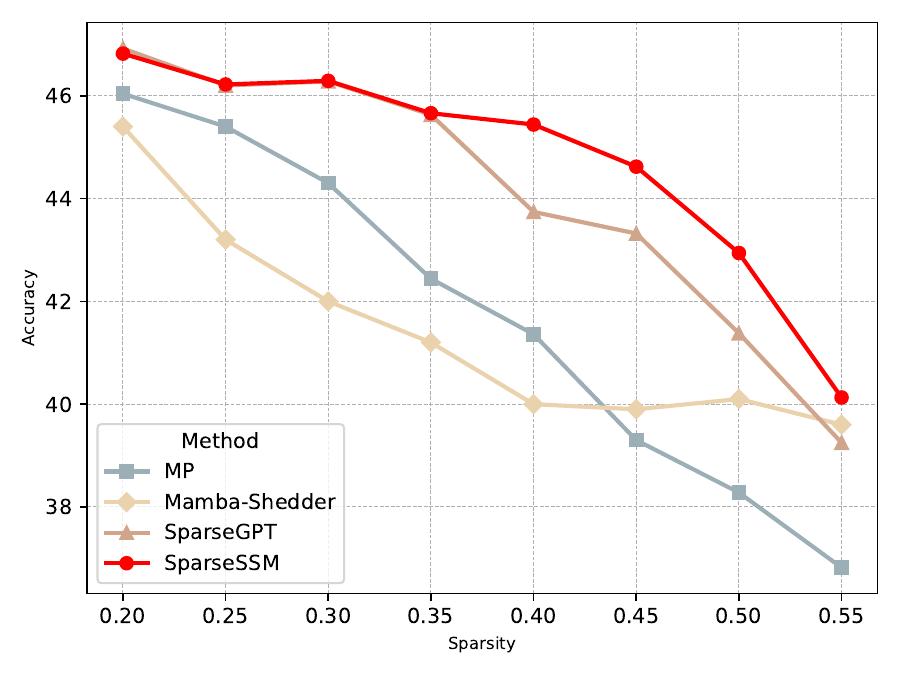}
  \caption{ Performance of the full Mamba architecture at multiple sparsity levels by measuring zero-shot task accuracy and Wikitext perplexity}
  \label{fig:all_sp}
\end{wrapfigure}

We then apply one‐shot unstructured pruning across all trainable weights except the input embedding and output head.
In this setting, each model typically incorporates an \texttt{nn.Conv1d} layer for feature preprocessing, \texttt{in\_proj} and \texttt{out\_proj} linear layers for dimensionality transformation, and—immediately before the selective scan operation, a learnable \texttt{x\_proj} mapping that produces the parameters $\Delta, B, C$, concurrently, the temporal stride parameter $\Delta$ is reparameterized via \texttt{dt\_proj}.
Empirical analysis reveals that these modules exhibit markedly heterogeneous pruning tolerances: pruning of the \texttt{in\_proj} and \texttt{out\_proj} layers induces substantially larger degradations than other linear modules, detailed comparison results are in Appendix~\ref{sec:details_different_modules}.

However, when we jointly prune both the SSM modules and the FFN branches, our proposed method \modelname\ outperforms all baselines, achieving lower perplexity and higher zero‐shot accuracy across every model scale, as shown in Table~\ref{tab:all-0.5}. 
Fig.~\ref{fig:all_sp} further illustrates our method’s performance across multiple sparsity levels. 
As demonstrated, for each downstream task, the pruned models exhibit consistent improvements, with gains becoming especially pronounced under higher sparsity regimes.

\begin{table}[!htbp]
\centering
\caption{Performance analysis for one-shot unstructured pruning of the whole Mamba models (130M $\sim$ 1.4B) at $50\%$ sparsity. 
Here, $\downarrow$ lower metrics reflect better outcomes, and $\uparrow$ denotes higher metrics reflect better outcomes.}
\vspace{1mm}
\resizebox{0.95\columnwidth}{!}{%
\renewcommand{\arraystretch}{1.3}
\begin{tabular}{c|c|*{9}{W{c}{3em}}}
\Xhline{3\arrayrulewidth}
Model & Method  & Wiki. $\downarrow$ & PTB $\downarrow$ & C4 $\downarrow$ 
      & OBQA $\uparrow$ & PIQA $\uparrow$ & ARC-e $\uparrow$ & ARC-c $\uparrow$
      & WinoG $\uparrow$ & Avg. $\uparrow$ \\ \hline
\multirow{5}{*}{Mamba-130M} 
  & Dense          & 20.60 & 32.75 & 25.66 & 28.60 & 63.28 & 48.02 & 24.40 & 52.5 & 43.36 \\ 
  & MP \citep{magnitude}            & 7.2e13 & 1.6e13  & 3.9e12 & \underline{27.00} & 50.82 & 25.88 & \textbf{27.47} & 49.41 & 36.12 \\ 
  & Mamba-Shedder \citep{mamba-shedder}  & \underline{364.8} & \underline{476.9} & \underline{231.4} & 24.4 & \underline{54.46} & \underline{34.68} & 22.95 & 49.41 & \underline{37.18} \\
  & SparseGPT \citep{sparsegpt}     & 6.2e7 & 1.8e7 & 7.5e5 & \textbf{27.40} & 53.42 & 28.24 & \underline{23.63} & \textbf{52.49} & 37.04 \\
  & \modelname     & \textbf{59.17} & \textbf{100.9} & \textbf{68.60} 
                   & 25.80 & \textbf{58.54} & \textbf{39.31} & 23.46
                   & \underline{49.80} & \textbf{39.38} \\ \hline 
\multirow{5}{*}{Mamba-370M} 
  & Dense          & 14.32 & 23.46 & 19.37 & 31.00 & 68.34 & 54.97 & 27.90 & 55.25 & 47.49 \\ 
  & MP \citep{magnitude}             & 3.7e10 & 8.3e10 & 3.1e10 & \textbf{30.40} & 55.71 & 31.40 & 24.15 & 49.96 & 38.32 \\ 
  & Mamba-Shedder \citep{mamba-shedder}  & \underline{192.3} & \underline{196.2} & \underline{120.1} & 27.60 & 57.94 & 39.90 & 22.78 & \textbf{52.33} & 40.11 \\
  & SparseGPT \citep{sparsegpt}     & 3.8e4 & 3.8e6 & 7717 & 28.60 & \underline{59.14} & \underline{42.00} & \textbf{24.83} & \textbf{52.33} & \underline{41.38} \\
  & \modelname     & \textbf{36.89} & \textbf{60.74} & \textbf{49.00} 
                   & \underline{30.00} & \textbf{62.24} & \textbf{45.96} & \underline{24.49} & \underline{52.01} & \textbf{42.94} \\ \hline 
\multirow{5}{*}{Mamba-790M} 
  & Dense          & 11.96 & 18.45 & 16.62 & 33.80 & 72.63 & 61.07 & 29.44 & 56.27 & 50.64 \\ 
  & MP \citep{magnitude}            & 6.6e57 & 4.5e53 & 2.5e46 & 26.40 & 54.24 & 28.11 & \textbf{25.60} & 48.54 & 36.58 \\ 
  & Mamba-Shedder \citep{mamba-shedder}  & \underline{121.9} & \underline{150.9} & \underline{112.7} & 25.80 & 57.78 & 38.47 & 22.61 & 49.64 & 38.86 \\
  & SparseGPT \citep{sparsegpt}     & 201.1 & 361.35 & 156.31 & \textbf{29.60} & \underline{62.62} & \underline{46.17} & 25.00 & \underline{51.86} & \underline{43.05} \\
  & \modelname     & \textbf{22.76} & \textbf{37.65} & \textbf{31.21} 
                   & \underline{29.00} & \textbf{64.58} & \textbf{50.04} & \underline{25.51} & \textbf{53.67} & \textbf{44.56} \\ \hline 
\multirow{5}{*}{Mamba-1.4B} 
  & Dense          & 10.75 & 17.05 & 15.17 & 36.40 & 73.88 & 65.57 & 32.85 & 61.17 & 53.98 \\
  & MP \citep{magnitude}            & 468.4 & 743.2 & 198.5 & 30.60 & \textbf{67.95} & \underline{53.28} & 24.06 & 52.49 & 45.68 \\ 
  & Mamba-Shedder \citep{mamba-shedder}   & 83.70 & 122.3 & 81.35 & 24.20 & 59.41 & 42.21 & 22.87 & 51.70 & 40.08 \\
  & SparseGPT \citep{sparsegpt}     & \underline{59.16} & \underline{95.14} & \underline{55.09} & \textbf{31.40} & \underline{67.74} & 53.03 & \underline{24.66} & \underline{54.70} & \underline{46.30} \\
  & \modelname     & \textbf{19.65} & \textbf{45.91} & \textbf{25.81} 
                   & \underline{30.80} & 66.10 & \textbf{56.06} & \textbf{26.62} & \textbf{56.59} & \textbf{47.24} \\
\Xhline{3\arrayrulewidth}
\end{tabular}%
}
\label{tab:all-0.5}
\end{table}

\subsection{Results of Semi-Structure and Structure Sparsity Extension} \label{sec:ex_structure}

Our approach admits a straightforward extension to $N:M$ and fully structured pruning. In fact, during unstructured pruning experiments, we observed that the pruned entries in the parameter $A_{log}$ overwhelmingly cluster within particular columns.
Certain hidden state channels in the state space model exhibit markedly higher redundancy. This empirical finding underpins the strong performance of our structured pruning scheme. 

Table~\ref{tab:exp-nm} reports results on the Mamba-370M model under 2:4 and 4:8 sparsity patterns. At the same overall sparsity, our method delivers smaller performance degradation in $N:M$ pruning.

To implement structured pruning, we target the second axis of $A$: we aggregate the importance of each column by computing its $L_1$ norm and then remove the least important columns. 
\begin{wraptable}{r}{0.5\textwidth}
  \centering
  \caption{Efficiency analysis of structured pruning at 50\% sparsity}
  \resizebox{0.85\linewidth}{!}{%
    \renewcommand{\arraystretch}{1.3}
    \begin{tabular}{c|>{\centering\arraybackslash}p{4cm}|>{\centering\arraybackslash}p{1cm}}
      \Xhline{3\arrayrulewidth}
      Sparsity & SSM inference time (ms) & Speedup \\ \hline
      Dense    & 2.766                    & /    \\
      50\%     & \textbf{1.608}           & \textbf{1.72$\times$} \\
      \Xhline{3\arrayrulewidth}
    \end{tabular}%
  }
  \label{tab:efficiency}
\end{wraptable}

Simultaneously, we resize the output dimension of the linear \texttt{x\_proj} layer to preserve tensor compatibility.
As shown in Table~\ref{tab:exp-structure}, this structured pruning on Mamba-370M induces only negligible accuracy loss at 50\% sparsity without any fine‐tuning, while accelerating the SSM module by a factor of $1.72 \times$. Table~\ref{tab:efficiency} shows the detailed inference time of SSM modules.

\vspace{-1.5mm}
\begin{table}[!htbp]
\centering
\caption{ Performance analysis for one-shot pruning of the SSM module in Mamba-370M at $2:4$ and $4:8$ sparsity patterns. 
}
\vspace{1mm}
\resizebox{0.95\columnwidth}{!}{%
\renewcommand{\arraystretch}{1.3}
\begin{tabular}{c|c|*{9}{W{c}{3em}}}
\Xhline{3\arrayrulewidth}
Sparsity & Method  & Wiki. $\downarrow$ & PTB $\downarrow$ & C4 $\downarrow$ 
      & OBQA $\uparrow$ & PIQA $\uparrow$ & ARC-e $\uparrow$ & ARC-c $\uparrow$
      & WinoG $\uparrow$ & Avg. $\uparrow$ \\ \hline

\multirow{2}{*}{2 : 4} 
  & MP             & 77.20 & 135.9 & 59.74 & 25.60 & 56.80 & 34.85 & 21.67 & 51.07 & 38.00 \\ 
  & \modelname     & \textbf{17.07} & \textbf{28.64} & \textbf{22.37} 
                   & \textbf{29.80} & \textbf{60.77} & \textbf{43.27} & \textbf{23.63}
                   & \textbf{54.14} & \textbf{42.32} \\ \hline 

\multirow{2}{*}{4 : 8} 
  & MP             & 81.56 & 148.25 & 63.76 & 26.00 & 55.44 & 37.54 & 21.84 & 50.12 & 38.19 \\ 
  & \modelname     & \textbf{16.77} & \textbf{28.27} & \textbf{22.16} 
                   & \textbf{31.40} & \textbf{64.47} & \textbf{46.00} & \textbf{25.77} & \textbf{51.14} & \textbf{43.76} \\ \hline

\Xhline{3\arrayrulewidth}
\end{tabular}%
}

\label{tab:exp-nm}
\end{table}

\vspace{-2mm}
\begin{table}[!htbp]
\centering
\caption{Performance analysis for one-shot structured pruning of the SSM module in Mamba-370M. 
}
\vspace{1mm}
\resizebox{0.95\columnwidth}{!}{%
\renewcommand{\arraystretch}{1.3}
\begin{tabular}{c|c|*{9}{W{c}{3em}}}
\Xhline{3\arrayrulewidth}
Sparsity & Method  & Wiki. $\downarrow$ & PTB $\downarrow$ & C4 $\downarrow$ 
      & OBQA $\uparrow$ & PIQA $\uparrow$ & ARC-e $\uparrow$ & ARC-c $\uparrow$
      & WinoG $\uparrow$ & Avg. $\uparrow$ \\ \hline

\multirow{2}{*}{25\%} 
  & MP             & 35.27 & 71.12  & 33.85 & 27.40 & 60.94 & 43.43 & 24.91 & 50.75 & 41.49 \\ 
  & \modelname     & \textbf{15.22} & \textbf{24.80} & \textbf{20.38} 
                   & \textbf{30.60} & \textbf{68.44} & \textbf{53.66} & \textbf{27.30} & \textbf{54.22} & \textbf{46.85} \\ \hline 

\multirow{2}{*}{50\%} 
  & MP             & 117.0 & 162.7 & 66.74 & 26.00 & 55.82 & 35.69 & 23.29 & 49.09 & 37.98 \\ 
  & \modelname     & \textbf{18.13} & \textbf{28.82} & \textbf{22.65} 
                   & \textbf{30.40} & \textbf{67.68} & \textbf{53.53} & \textbf{27.30} & \textbf{52.64} & \textbf{46.31} \\ \hline

\Xhline{3\arrayrulewidth}
\end{tabular}%
}

\label{tab:exp-structure}
\end{table}

\subsection{Ablation Study} \label{sec:ex_ablation}

We conduct a systematic ablation study on the key components of \modelname\ to isolate their contributions to pruning efficacy.
In particular, we find that accurate Hessian matrix estimation is instrumental to our method’s superior performance, while incorporating a temporal pruning‐frequency metric yields additional gains. 
As shown in Table~\ref{tab:timestep}, our full strategy significantly outperforms a simpler baseline that applies our Hessian estimate via a simple $L_2$ norm aggregation over time steps.

\begin{table}[!htbp]
\centering
\caption{Performance analysis for different methods of time steps aggregation. 
We conduct our experiments on Mamba-370M at multiple sparsities.
}
\vspace{1mm}
\resizebox{0.95\columnwidth}{!}{%
\renewcommand{\arraystretch}{1.3}
\begin{tabular}{c|c|*{9}{W{c}{3em}}}
\Xhline{3\arrayrulewidth}
Sparsity & Method  & Wiki. $\downarrow$ & PTB $\downarrow$ & C4 $\downarrow$ 
      & OBQA $\uparrow$ & PIQA $\uparrow$ & ARC-e $\uparrow$ & ARC-c $\uparrow$
      & WinoG $\uparrow$ & Avg. $\uparrow$ \\ \hline

\multirow{2}{*}{50\%} 
  & L2             & 81.22 & 183.9 & 51.49 & 30.40 & 66.10 & 50.51 & 25.94 & 53.28 & 45.24 \\ 
  & \modelname     & \textbf{19.27} & \textbf{31.05} & \textbf{24.72} 
                   & \textbf{32.80} & \textbf{69.64} & \textbf{54.21} & \textbf{27.05} & \textbf{53.67} & \textbf{47.47} \\ \hline

\multirow{2}{*}{60\%} 
  & L2             & 108.1 & 242.0 & 62.14 & 28.40 & 62.89 & 46.84 & 25.68 & 51.54 & 43.07 \\ 
  & \modelname     & \textbf{22.65} & \textbf{38.37} & \textbf{28.10} 
                   & \textbf{31.80} & \textbf{66.76} & \textbf{49.45} & \textbf{27.82} & \textbf{52.80} & \textbf{45.73} \\ \hline 

\multirow{2}{*}{70\%} 
  & L2             & 186.3 & 372.1 & 82.61 & 30.20 & 61.04 & 44.03 & 23.21 & 51.30 & 41.96 \\ 
  & \modelname     & \textbf{28.28} & \textbf{44.93} & \textbf{33.28} 
                   & \textbf{27.00} & \textbf{65.56} & \textbf{49.03} & \textbf{24.06} & \textbf{53.35} & \textbf{43.80} \\ \hline 

\Xhline{3\arrayrulewidth}
\end{tabular}%
}

\label{tab:timestep}
\end{table}

\vspace{1.5mm}

\begin{figure}[ht]
  \centering
  \begin{subfigure}[b]{0.48\textwidth}
    \centering
    \includegraphics[trim=0.2cm 0.1cm 0.2cm 0.1cm,
    clip, width=\linewidth]{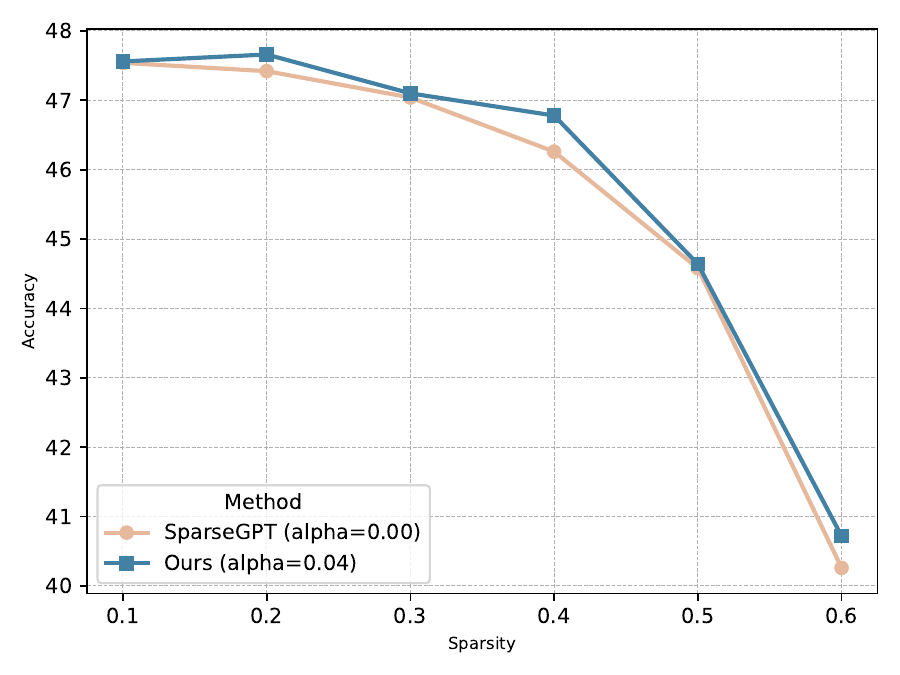}
    \label{fig:ffn-A}
  \end{subfigure} 
  \begin{subfigure}[b]{0.48\textwidth}
    \centering
    \includegraphics[trim=0.2cm 0.1cm 0.2cm 0.1cm,
    clip, width=\linewidth]{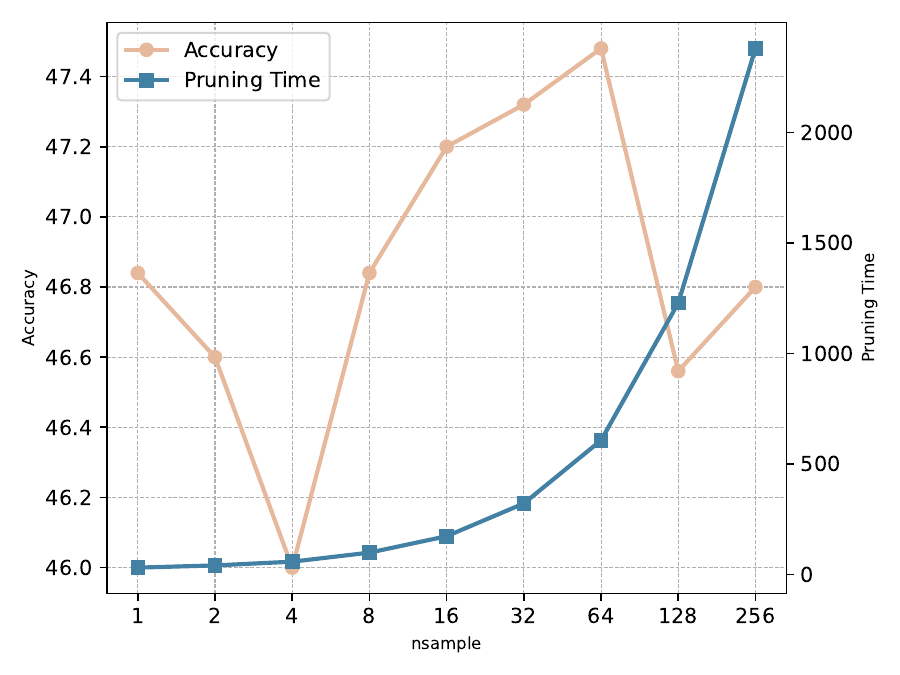}
    \label{fig:ffn-B}
  \end{subfigure}
  \caption{ Effects of calibration sample size and sparsity interval. \textbf{(Left)} shows a performance analysis of pruning the FFN components of Mamba-370M under varying sparsity settings, while \textbf{(Right)} shows a performance analysis and pruning efficiency analysis of pruning the SSM modules as the calibration sample size is varied.}
  \label{fig:albation}
\end{figure}

We further assess the effects of sensitivity pruning width and calibration data volume on final results.
We change the super parameter $\alpha$ that controls pruning width and $N_{sample}$ that controls calibration data volume.
Our experiments reveal that selecting an appropriately sized width parameter substantially improves pruning outcomes in the FFN components, surpassing the performance of SparseGPT.
As for calibration data, we observe that fewer than 16 samples degrade the performance of the pruned model. 
However, a sampling count of 64 strikes the best trade‐off between pruning quality and computational efficiency.

\section{Conclusion} \label{sec:conclusion}

In this work, we introduce {\modelname}, a one‐shot, training‐free unstructured pruning framework that extends the classic OBS paradigm to selective state‐space modules in Mamba-based LLMs.
By incorporating time‐sharing parameter saliency and explicitly accounting for the discretization of the state‐transition matrix, our layer‐wise algorithm computes local second‐order importance scores and reconstructs remaining weights to minimize output error. 
Furthermore, our module sensitivity analysis reveals distinct pruning tolerances between input and output projections, offering new insights into redundancy within state‐space architectures.
Our results establish that state‐space LLMs like Mamba can be compressed as effectively as their Transformer counterparts via principled, OBS‐guided pruning, paving the way for more efficient deployment within resource‐restricted contexts.
In future work, we plan to further extend \modelname\ to structured pruning of the entire Mamba architecture. We also aim to generalize our approach to other time‐varying architectures and investigate hardware‐aware optimizations that further accelerate sparse state‐space inference.

\newpage
\bibliographystyle{unsrt}
\bibliography{references}

\begin{thebibliography}{10}

\bibitem{LLaMA}
Hugo Touvron, Thibaut Lavril, Gautier Izacard, Xavier Martinet, MarieAnne Lachaux, Timothe Lacroix, Baptiste Rozire, Naman Goyal, Eric Hambro, Faisal Azhar, Aurelien Rodriguez, Armand Joulin, Edouard Grave, and Guillaume Lample.
\newblock Llama: Open and efficient foundation language models.
\newblock {\em arXiv preprint arXiv:2302.13971}, 2023.

\bibitem{opt}
Susan Zhang, Stephen Roller, Naman Goyal, Mikel Artetxe, Moya Chen, Shuohui Chen, Christopher Dewan, Mona Diab, Xian Li, and Xi~Victoria~Lin et~al.
\newblock Opt: Open pre-trained transformer language models.
\newblock {\em arXiv preprint arXiv:2205.01068}, 2022.

\bibitem{bloom}
BigScience Workshop, :, Teven~Le Scao, Angela Fan, Christopher Akiki, Ellie Pavlick, Suzana Ilić, and Daniel~Hesslow et~al.
\newblock Bloom: A 176b-parameter open-access multilingual language model, 2023.

\bibitem{obd}
Y.~LeCun, J.~S. Denker, and S.~A. Solla.
\newblock Optimal brain damage.
\newblock In {\em NeurIPS}, 1990.

\bibitem{obs}
B.~Hassibi and D.~G. Stork.
\newblock Second order derivatives for network pruning: Optimal brain surgeon.
\newblock In {\em NeurIPS}, 1993.

\bibitem{deep-compression}
Song Han, Huizi Mao, and William~J Dally.
\newblock Deep compression: Compressing deep neural networks with pruning, trained quantization and huffman coding.
\newblock In {\em ICLR}, 2016.

\bibitem{llm-pruner}
Xinyin Ma, Gongfan Fang, and Xinchao Wang.
\newblock Llm-pruner: On the structural pruning of large language models.
\newblock In {\em NeurIPS}, 2023.

\bibitem{sparsegpt}
Elias Frantar and Dan Alistarh.
\newblock {SparseGPT}: Massive language models can be accurately pruned in one-shot.
\newblock In {\em ICML}, 2023.

\bibitem{magnitude}
Song Han, Jeff Pool, John Tran, and William~J Dally.
\newblock Learning both weights and connections for efficient neural network.
\newblock In {\em NeurIPS}, 2015.

\bibitem{pruning-filters}
Hao Li, Asim Kadav, Igor Durdanovic, Hanan Samet, and Hans~Peter Graf.
\newblock Pruning filters for efficient convnets.
\newblock In {\em ICLR}, 2017.

\bibitem{channel-pruning}
Yihui He, Xiangyu Zhang, and Jian Sun.
\newblock Channel pruning for accelerating very deep neural networks.
\newblock In {\em ICCV}, 2017.

\bibitem{snip}
Namhoon Lee, Thalaiyasingam Ajanthan, and Philip Torr.
\newblock Snip: Single-shot network pruning based on connection sensitivity.
\newblock In {\em ICLR}, 2019.

\bibitem{synflow}
Hidenori Tanaka, Daniel Kunin, Daniel~L Yamins, and Surya Ganguli.
\newblock Pruning neural networks without any data by iteratively conserving synaptic flow.
\newblock In {\em NeurIPS}, 2020.

\bibitem{obc}
Elias Frantar and Dan Alistarh.
\newblock Optimal brain compression: A framework for accurate post-training quantization and pruning.
\newblock In {\em ICML}, 2022.

\bibitem{wanda}
Mingjie Sun, Zhuang Liu, Anna Bair, and J.~Zico Kolter.
\newblock A simple and effective pruning approach for large language models.
\newblock {\em arXiv preprint arXiv:2306.11695}, 2023.

\bibitem{ALPS}
Xiang Meng, Kayhan Behdin, Haoyue Wang, and Rahul Mazumder.
\newblock Alps: Improved optimization for highly sparse one-shot pruning for large language models.
\newblock In {\em NeurIPS}, 2024.

\bibitem{mamba}
Albert Gu and Tri Dao.
\newblock Mamba: Linear-time sequence modeling with selective state spaces.
\newblock {\em arXiv preprint arXiv:2312.00752}, 2023.

\bibitem{mamba2}
Tri Dao and Albert Gu.
\newblock Transformers are {SSM}s: Generalized models and efficient algorithms through structured state space duality.
\newblock In {\em ICML}, 2024.

\bibitem{falcon-mamba}
J.~Zuo, M.~Velikanov, D.~E. Rhaiem, et~al.
\newblock Falcon mamba: The first competitive attention-free 7b language model.
\newblock {\em arXiv preprint arXiv:2410.05355}, 2024.

\bibitem{transformer}
Ashish Vaswani, Noam Shazeer, Niki Parmar, Jakob Uszkoreit, Llion Jones, Aidan~N Gomez, {\L}ukasz Kaiser, and Illia Polosukhin.
\newblock Attention is all you need.
\newblock In {\em NeurIPS}, 2017.

\bibitem{hippo}
Albert Gu, Tri Dao, Stefano Ermon, Atri Rudra, and Christopher Re.
\newblock Hippo: Recurrent memory with optimal polynomial projections.
\newblock In {\em NeurIPS}, 2020.

\bibitem{s4}
Albert Gu, Karan Goel, and Christopher R\'e.
\newblock Efficiently modeling long sequences with structured state spaces.
\newblock In {\em ICLR}, 2022.

\bibitem{s5}
Jimmy~T.H. Smith, Andrew Warrington, and Scott Linderman.
\newblock Simplified state space layers for sequence modeling.
\newblock In {\em ICLR}, 2023.

\bibitem{jamba}
Opher Lieber, Barak Lenz, Hofit Bata, Gal Cohen, Jhonathan Osin, Itay Dalmedigos, Erez Safahi, Shaked Meirom, Yonatan Belinkov, Shai Shalev-Shwartz, Omri Abend, Raz Alon, Tomer Asida, Amir Bergman, Roman Glozman, Michael Gokhman, Avashalom Manevich, Nir Ratner, Noam Rozen, Erez Shwartz, Mor Zusman, and Yoav Shoham.
\newblock Jamba: A hybrid transformer-mamba language model.
\newblock In {\em ICLR}, 2024.

\bibitem{zamba}
P.~Glorioso, Q.~Anthony, Y.~Tokpanov, et~al.
\newblock The zamba2 suite: Technical report.
\newblock {\em arXiv preprint arXiv:2411.15242}, 2024.

\bibitem{simba}
Badri~N Patro and Vijay~S Agneeswaran.
\newblock Simba: Simplified mamba-based architecture for vision and multivariate time series.
\newblock {\em arXiv preprint arXiv:2403.15360}, 2024.

\bibitem{hymba}
Xin Dong, Yonggan Fu, Shizhe Diao, Wonmin Byeon, Zijia Chen, Ameya~Sunil Mahabaleshwarkar, Shih-Yang Liu, Matthijs Van~Keirsbilck, Min-Hung Chen, Yoshi Suhara, Yingyan Lin, Jan Kautz, and Pavlo Molchanov.
\newblock Hymba: A hybrid-head architecture for small language models.
\newblock {\em arXiv preprint arXiv:2411.13676}, 2024.

\bibitem{woodfisher}
Sidak~Pal Singh and Dan Alistarh.
\newblock Woodfisher: Efficient second-order approximations for model compression.
\newblock In {\em NeurIPS}, 2020.

\bibitem{chita}
Riade Benbaki, Wenyu Chen, Xiang Meng, Hussein Hazimeh, Natalia Ponomareva, Zhe Zhao, and Rahul Mazumder.
\newblock Fast as chita: Neural network pruning with combinatorial optimization.
\newblock In {\em ICML}, 2023.

\bibitem{bert}
Jacob Devlin, Ming-Wei Chang, Kenton Lee, and Kristina Toutanova.
\newblock Bert: Pre-training of deep bidirectional transformers for language understanding.
\newblock In {\em NAACL}, 2019.

\bibitem{llm-surgeon}
Tycho F.~A. van~der Ouderaa, Markus Nagel, Mart van Baalen, Yuki~M. Asano, and Tijmen Blankevoort.
\newblock The llm surgeon.
\newblock In {\em ICLR}, 2024.

\bibitem{darwinlm}
Shengkun Tang, Oliver Sieberling, Eldar Kurtic, Zhiqiang Shen, and Dan Alistarh.
\newblock Darwinlm: Evolutionary structured pruning of large language models.
\newblock {\em arXiv preprint arXiv:2502.07780}, 2025.

\bibitem{slimgpt}
Gui Ling, Ziyang Wang, Yuliang Yan, and Qingwen Liu.
\newblock Slimgpt: Layer-wise structured pruning for large language models.
\newblock In {\em NeurIPS}, 2024.

\bibitem{structured-obs}
Jiateng Wei, Quan Lu, Ning Jiang, Siqi Li, Jingyang Xiang, Jun Chen, and Yong Liu.
\newblock Structured optimal brain pruning for large language models.
\newblock In {\em NeurIPS}, 2024.

\bibitem{i-obs}
Diyuan Wu, Ionut-Vlad Modoranu, Mher Safaryan, Denis Kuznedelev, and Dan Alistarh.
\newblock The iterative optimal brain surgeon: Faster sparse recovery by leveraging second-order information.
\newblock In {\em NeurIPS}, 2024.

\bibitem{c-obs}
Xin Yu, Thiago Serra, Srikumar Ramalingam, and Shandian Zhe.
\newblock The combinatorial brain surgeon: Pruning weights that cancel one another in neural networks.
\newblock In {\em ICML}, 2022.

\bibitem{kwak2024layer}
Minsunu Kwak, Seungrok Moon, Joohwan Ko, and POOGYEON PARK.
\newblock Layer-adaptive state pruning for deep state space models.
\newblock In {\em NeurIPS}, 2024.

\bibitem{mamba-shedder}
J.~Pablo Mu{\~n}oz, Jinjie Yuan, and Nilesh Jain.
\newblock Mamba-shedder: Post-transformer compression for efficient selective structured state space models.
\newblock In {\em NAACL}, 2025.

\bibitem{mambap}
T.~Ghattas, M.~Hassid, and R.~Schwartz.
\newblock On pruning state-space llms.
\newblock {\em arXiv preprint arXiv:2502.18886}, 2025.

\bibitem{hybridp}
A.~Taghibakhshi, S.~T. Sreenivas, S.~Muralidharan, et~al.
\newblock Efficient hybrid language model compression through group-aware ssm pruning.
\newblock {\em arXiv preprint arXiv:2504.11409}, 2025.

\bibitem{bptt}
P.J. Werbos.
\newblock Backpropagation through time: what it does and how to do it.
\newblock {\em Proceedings of the IEEE}, 1990.

\bibitem{sensitive}
Hang Shao, Bei Liu, and Yanmin Qian.
\newblock One-shot sensitivity-aware mixed sparsity pruning for large language models.
\newblock In {\em ICASSP}, 2024.

\bibitem{ppl}
{Hugging Face}.
\newblock Perplexity of fixed-length models.
\newblock 2022.

\bibitem{wikitext}
Stephen Merity, Caiming Xiong, James Bradbury, and Richard Socher.
\newblock Pointer sentinel mixture models.
\newblock In {\em ICLR}, 2017.

\bibitem{ptb}
Mitch Marcus, Grace Kim, Mary~Ann Marcinkiewicz, Robert MacIntyre, Ann Bies, Mark Ferguson, Karen Katz, and Britta Schasberger.
\newblock The penn treebank: Annotating predicate argument structure.
\newblock In {\em Human Language Technology: Proceedings of a Workshop held at Plainsboro, New Jersey, March 8-11, 1994}, 1994.

\bibitem{c4}
Colin Raffel, Noam Shazeer, Adam Roberts, Katherine Lee, Sharan Narang, Michael Matena, Yanqi Zhou, Wei Li, and Peter~J Liu.
\newblock Exploring the limits of transfer learning with a unified text-to-text transformer.
\newblock {\em JMLR}, 21(140):1--67, 2020.

\bibitem{piqa}
Yonatan Bisk, Rowan Zellers, Jianfeng Gao, Yejin Choi, et~al.
\newblock Piqa: Reasoning about physical commonsense in natural language.
\newblock In {\em AAAI}, 2020.

\bibitem{openbookqa}
Todor Mihaylov, Peter Clark, Tushar Khot, and Ashish Sabharwal.
\newblock Can a suit of armor conduct electricity? a new dataset for open book question answering.
\newblock In {\em EMNLP}, 2018.

\bibitem{winogrande}
Keisuke Sakaguchi, Ronan~Le Bras, Chandra Bhagavatula, and Yejin Choi.
\newblock Winogrande: An adversarial winograd schema challenge at scale.
\newblock {\em arXiv preprint arXiv:1907.10641}, 2019.

\bibitem{ai2}
Peter Clark, Isaac Cowhey, Oren Etzioni, Tushar Khot, Ashish Sabharwal, Carissa Schoenick, and Oyvind Tafjord.
\newblock Think you have solved question answering? try arc, the ai2 reasoning challenge.
\newblock {\em arXiv preprint arXiv:1803.05457}, 2018.

\bibitem{mamba_minimal}
John Ma.
\newblock {mamba-minimal}: A minimal pytorch implementation of mamba.
\newblock \url{https://github.com/johnma2006/mamba-minimal}.

\bibitem{lstm}
Sepp Hochreiter and J\"{u}rgen Schmidhuber.
\newblock Long short-term memory.
\newblock {\em Neural Computation}, 9(8):1735–1780, 1997.

\end{thebibliography}

\newpage 
\appendix

\section{Proofs of Theorem \textcolor{red}{1}} \label{sec:proof-obs}

\noindent\textbf{\textit{Proof.}} 
We begin at a trained network’s parameters, where $A_{\log}$ is near a local minimum of the loss $L$. 
In this setting, small perturbations of the parameters cause a loss increase dominated by the quadratic term of the second-order Taylor expansion. 
\begin{lemma}[OBS Parameter Importance]\label{lem:obs-importance}
Under the second‐order OBS pruning framework, let $\mathbf{H = \nabla^2 L(\theta)}$ denote the Hessian of the loss \(L\) with respect to the full parameter vector \(\theta\).  Then the saliency of the individual parameter \(\mathbf{A_{\log,d,n}}\) is given by
\begin{equation}
    I^{\log}_{d,n}
    \;=\;
    \frac{\bigl(A_{\log,d,n}\bigr)^2}
         {2\,\bigl[H^{-1}\bigr]_{(d,n),(d,n)}}
    \;=\;
    \frac{1}{2}\;H_{(d,n),(d,n)}\;\bigl(A_{\log,d,n}\bigr)^2.
    \label{eq:I_log}
\end{equation}
\end{lemma}

Returning to the proof of the main theorem, to calculate the saliency, we now calculate the Hessian matrix $H$ by propagating derivatives through the SSM dynamics.
Using the chain rule, the first derivative of the loss with respect to $A_{\log,d,n}$ is:
\begin{equation}
    \frac{\partial L}{\partial A_{\log,d,n}} \;=\; \sum_{b,i} \frac{\partial L}{\partial h_{b,i,d,n}}\;\frac{\partial h_{b,i,d,n}}{\partial A_{\log,d,n}}\,.
\end{equation}
For brevity, let us denote $w_{b,i,d,n} = e^{A_{d,n},\delta_{b,i,d}}$, then we have
\begin{equation}
\begin{aligned}
\frac{\partial h_{b,i,d,n}}{\partial A_{\log,d,n}}
&=
\frac{\partial w_{b,i,d,n}}{\partial A_{d,n}}
\;\frac{\partial A_{d,n}}{\partial A_{\log,d,n}}
\;h_{b,i-1,d,n}
\\
&=
A_{d,n}\,\delta_{b,i,d}\,w_{b,i,d,n}\;h_{b,i-1,d,n}\,.
\label{eq:h/alog}
\end{aligned}
\end{equation}
Substituting Eq.~\eqref{eq:h/alog} into the expression for $\partial L \ /\ \partial A_{\log,d,n}$:
\begin{equation}
    \frac{\partial L}{\partial A_{\log,d,n}} \;=\; \sum_{b,i} \frac{\partial L}{\partial h_{b,i,d,n}}\;\Big(A_{d,n}\,\delta_{b,i,d}\,e^{A_{d,n}\delta_{b,i,d}}\;h_{b,i-1,d,n}\Big)\,. 
\end{equation}
Thus, let us differentiate again to get the second derivative $\partial^2 L/\partial A_{\log,d,n}^2$. Differentiating the above expression with respect to $A_{\log,d,n}$ yields:
\begin{equation}
    \frac{\partial^2 L}{\partial A_{\log,d,n}^2} \;=\; \sum_{b,i} \frac{\partial}{\partial A_{\log,d,n}}\Big[\frac{\partial L}{\partial h_{b,i,d,n}}\;A_{d,n}\,\delta_{b,i,d}\,e^{A_{d,n}\delta_{b,i,d}}\;h_{b,i-1,d,n}\Big]\,.
\end{equation}
Differentiating $\frac{\partial L}{\partial h_{b,i,d,n}}$ (the backpropagated gradient) with respect to $A_{\log,d,n}$ corresponds to third-order effects.
Thus, we approximate:
\begin{align}
\frac{\partial^{2}L}{\partial A_{\log,d,n}^{2}}
&\approx
\sum_{b,i}\frac{\partial L}{\partial h_{b,i,d,n}}
          \frac{\partial}{\partial A_{\log,d,n}}
          \Bigl(A_{d,n}\,\delta_{b,i,d}\,e^{A_{d,n}\delta_{b,i,d}}\,
          h_{b,i-1,d,n}\Bigr)
          \displaybreak[2]\\
&=
\sum_{b,i}\frac{\partial L}{\partial h_{b,i,d,n}}
   \Bigl[
       \frac{\partial A_{d,n}}{\partial A_{\log,d,n}}\,
       \delta_{b,i,d}\,e^{A_{d,n}\delta_{b,i,d}}
       + A_{d,n}\,
         \frac{\partial}{\partial A_{\log,d,n}}
         \bigl(\delta_{b,i,d}\,e^{A_{d,n}\delta_{b,i,d}}\bigr)
   \Bigr]
   h_{b,i-1,d,n}
   \displaybreak[2]\\
&=
\sum_{b,i}\frac{\partial L}{\partial h_{b,i,d,n}}
   \Bigl[
       A_{d,n}\,\delta_{b,i,d}\,e^{A_{d,n}\delta_{b,i,d}}
       + A_{d,n}^{2}\,\delta_{b,i,d}^{2}\,e^{A_{d,n}\delta_{b,i,d}}
   \Bigr]
   h_{b,i-1,d,n}
   \displaybreak[2]\\
&=
\sum_{b,i}
   \underbrace{\frac{\partial L}{\partial h_{b,i,d,n}}\,
               A_{d,n}^{2}\,\delta_{b,i,d}^{2}\,
               e^{A_{d,n}\delta_{b,i,d}}\,
               h_{b,i-1,d,n}}_{\text{second-order term at time }i}\,.
\end{align}
The factor $\partial L/\partial h_{b,i,d,n}$ can now be interpreted as the first-order loss gradient at that state, for example,
\begin{equation}
    \frac{\partial^2 L}{\partial (A_{d,n} h_{b,i-1,d,n})^2} \;=\;  \frac{\partial L}{\partial h_{b,i,d,n}}\Big/ h_{b,i-1,d,n}\,.
\end{equation}
We can then rewrite the above Hessian approximation as:
\begin{equation}
    H_{(d,n),(d,n)} \;=\; \frac{\partial^2 L}{\partial A_{\log,d,n}^2} \;\approx\; \kappa \sum_{b,i} A_{d,n}^2\,\delta_{b,i,d}^2\,e^{2\,A_{d,n}\delta_{b,i,d}}\;h_{b,i-1,d,n}^2\,.
\end{equation}
Substituting this in Eq.~\eqref{eq:I_log}, we obtain:
\begin{equation}
    I^{\log}_{d,n} \;=\; H_{(d,n),(d,n)}\,\big(A_{\log,d,n}\big)^2 \;\approx\; \kappa\,A_{d,n}^2\,\big(A_{\log,d,n}\big)^2 \sum_{b,i} \delta_{b,i,d}^2\,  e^{2\,A_{d,n}\delta_{b,i,d}}\;h_{b,i-1,d,n}^2\,.
    \label{eq:estimation}
\end{equation}
For Eq.~\eqref{eq:estimation}, we observe that the term $\delta_{b,i,d}^2$ is data-dependent but does not depend on the particular parameter being pruned.
The exponential term $A^2_{d, n}e^{2\,\delta_{b,i,d}A_{d,n}}$ varies much more slowly with $d,n$ than $h_{b,i-1,d,n}$ since $A_{d,n}$ is negative.
Therefore, we can write:
\begin{equation}
    I^{\log}_{d,n} \;=\; \kappa \sum_{b,i} \delta_{b,i,d}^2\, A_{d,n}^2e^{2\,\delta_{b,i,d}A_{d,n}}\;\times\; A_{\log,d,n}^2\;\sum_{b,i} h_{b,i-1,d,n}^2\,.
\end{equation}
This shows that up to a constant factor, the importance score for parameter $(d,n)$ is
\begin{equation}
    I^{\log}_{d,n} \;\propto\; A_{\log,d,n}^{2}\;\sum_{b,i}h_{b,i-1,d,n}^{2}\,. 
\end{equation}
The right-hand side is exactly the simple product stated in the theorem.
$\hfill \square$

\subsection{Proof of Lemma~\ref{lem:obs-importance}}

Let $\theta$ denote the vector of all parameters and $H = \nabla^2 L(\theta)$ the Hessian at the optimum. For a perturbation $\Delta \theta$, the Taylor expansion gives: 
\begin{equation}
    \Delta L \;\approx\; \frac{1}{2}\,\Delta \theta^T H\,\Delta \theta\,.
\end{equation}

In the SSM module, Over a small time increment $\delta_{b,i,d}$ at step $i$, the state update (solution of $\dot{h}=A_{d,n} h$) is:
\begin{equation}
    h_{b,i,d,n} \;=\; e^{A_{d,n}\,\delta_{b,i,d}}\;h_{b,i-1,d,n} \;+\; \Delta \bigl( B_{u} \bigr)_{i}\,,
\end{equation}
where $\Delta \bigl( B_{u} \bigr)_{i}$ is independent with parameter $A$.
The only way $A_{\log,d,n}$ affects the network’s forward pass is through this scalar multiplier $e^{A_{d,n},\delta_{b,i,d}}$ at each time step. 
Crucially, because $A$ is diagonal, each parameter $A_{\log,d,n}$ influences only its corresponding state dimension $d$ in SSM $n$, independently of other dimensions, which implies
\begin{equation}
    \frac{\partial^2 L}{\partial A_{\log,d,n}\partial A_{\log,d',n'}} \; = \; 0 \;,\;  \big(d',n'\big) \neq \big(d,n \big).
\end{equation}
Therefore, the Hessian matrix $H$ has the characteristic
\begin{equation}
    \bigl[H^{-1}\bigr]_{(d,n),(d,n)} = \frac{1}{H_{(d,n),(d,n)}}.
\end{equation}
where $H_{(d,n),(d,n)} = \partial^2 L \ /\  \partial A_{\log,d,n}^2$ is the Hessian’s diagonal entry for that parameter.
Combining with the classic OBS saliency definition $\varepsilon_{m} = w_{m}^{2} \ /\ [\mathbf{H}^{-1}]_{mm}\,$,then we define the OBS saliency of parameter $A_{log,d,n}$ as
\begin{equation}
    I^{\log}_{d,n} 
    \;=\; \frac{\big(A_{log,d,n} \big)^2}{2\bigl[ H \bigr]^{-1}_{(d,n),(d,n)}}
    \;=\; \frac{1}{2}H_{(d,n),(d,n)}\,\big(A_{\log,d,n}\big)^2\,.
\end{equation}
$\hfill \square$

\section{Experiments Details} \label{sec:details}

\subsection{Experiments Setup} \label{sec:details_setup}

We performed all experiments on a dedicated server using dual Intel Xeon Platinum 8457C processors (48 cores / 96 threads each), 512 GB of DDR5 memory, and eight NVIDIA GeForce RTX 4090 GPUs (24 GB each). 
We used the PyTorch library to implement the Mamba model and pruning methods for our experiments.

We based our implementation on the SparseGPT code framework \citep{sparsegpt}, performing pruning on a per-module basis by registering forward hooks to capture each module’s inputs during the forward pass. 
After pruning a given layer, we update its inputs to maintain correct activation propagation. For each pruned module, we remove the designated parameters to realize the prescribed sparsity.

In our Mamba implementation, we adopted the mamba-minimal \citep{mamba_minimal} code framework and loaded the official Mamba checkpoint \citep{mamba} for pretrained weights. 
To meet our experimental objectives, we introduced a small set of modifications to the mamba-minimal implementation.

\noindent\textbf{Hyperparameters.} \label{sec:hyperparameters}
For SSM-module pruning, we set $N_{sample} = 64$, which we found yields the best trade-off between pruning quality and computational cost.
In the FFN pruning stage, we chose $\alpha = 0.04$, implying that each FFN submodule is assigned a sparsity rate of
\begin{equation}
    S_{\mathrm{FFN},i} =
  \begin{cases}
  0.96 - p + \dfrac{0.08\,id}{N - 1}, & \text{if } i \in \{\mathtt{in\_proj},\,\mathtt{out\_proj}\},\\[6pt]
  S_{\mathrm{global}},                              & \text{otherwise,}
  \end{cases}
\end{equation}
where \(N\) is the total number of weights, and \(id\) is the sensitivity‑rank index of the given weight after sorting by Hessian‑trace importance.  It means that for the modules \texttt{in\_proj} and \texttt{out\_proj}, the allowable deviation interval \(\bigl[0.96 - S_{global},\;1.04 - S_{global}]\).
The remaining hyperparameters governed the logging and pruning module configuration.

\noindent\textbf{Implementation Details.} \label{sec:implementation}
Below, we summarize the precise configurations used for each selected baseline:
\begin{itemize}[noitemsep,topsep=0pt,parsep=0pt,partopsep=0pt, leftmargin=*]
    \item MP \citep{magnitude}:  The weight matrix of each module is sorted by absolute value, retaining the $top-k$ entries and zeroing out all others. For SSM modules, the same procedure is applied to the state-transition matrix $A$.
    \item Mamba-Shedder \citep{mamba-shedder}: We employed the authors’ published implementation and default settings, without fine-tuning. Since the authors built upon the official Mamba model implementation and introduced their own modifications, we reproduced this baseline by employing the Mamba model definition as provided by the authors.
    \item SparseGPT \citep{sparsegpt}: We extended the original \texttt{SparseGPT} framework to support Mamba pruning via two key adaptations: (1) when pruning \texttt{nn.Conv1d} modules, We applied the SparseGPT processing pipeline for \texttt{transformer.Conv1d} directly to the \texttt{nn.Conv1d} modules; and (2) when pruning the SSM parameter matrix $A$, we enable direct matrix-level pruning and use the hidden state $h$ as calibration data.
\end{itemize}

\subsection{Additional Experiments Results} \label{sec:details_additional_results}

\subsubsection{Pruning Efficiency Analysis} \label{sec:details_efficiency}

Our proposed method can prune Mamba-based large language models in an extremely short time. 
Specifically, thanks to our efficient Hessian matrix estimation method and fully parallelized implementation, the time required to compute pruning scores is virtually negligible; the primary time overhead instead stems from processing the calibration data.

\begin{table}[t]
  \centering
  \small
  \caption{Performance analysis of pruning time overhead. Specifically, we conduct experiments on multiple model variants and across different calibration‐data sample sizes.}
  \vspace{1mm}
  \label{tab:pruning_time}
  \renewcommand{\arraystretch}{1.3}
  \begin{tabularx}{0.75\textwidth}{
    >{\centering\arraybackslash}p{2.0cm}|
    >{\centering\arraybackslash}X|
    >{\centering\arraybackslash}p{1.8cm}|
    >{\centering\arraybackslash}X|
    >{\centering\arraybackslash}p{2.0cm}
  }
    \Xhline{3\arrayrulewidth}
    Model & Layers & Hidden size & Nsample & Pruning time \\ \hline
    \multirow{3}{*}{Mamba-130M} 
      & \multirow{3}{*}{24} 
      & \multirow{3}{*}{768} 
      & 32  & 164.4378\,s \\ 
      &                     &                      
      & 64  & 311.3634\,s \\ 
      &                     &                      
      & 128 & 624.6192\,s \\ \hline
    \multirow{3}{*}{Mamba-370M} 
      & \multirow{3}{*}{48} 
      & \multirow{3}{*}{1024} 
      & 32  & 319.0448\,s \\ 
      &                     &                      
      & 64  & 602.5500\,s \\ 
      &                     &                      
      & 128 & 1203.028\,s \\ \hline
    \multirow{3}{*}{Mamba-790M} 
      & \multirow{3}{*}{48} 
      & \multirow{3}{*}{1536} 
      & 32  & 326.1090\,s \\ 
      &                     &                      
      & 64  & 630.0898\,s \\ 
      &                     &                      
      & 128 & 1239.914\,s \\ \hline
    \multirow{3}{*}{Mamba-1.4B} 
      & \multirow{3}{*}{48} 
      & \multirow{3}{*}{2048} 
      & 32  & 348.4770\,s \\ 
      &                     &                      
      & 64  & 662.2011\,s \\ 
      &                     &                      
      & 128 & 1272.2396\,s \\ \Xhline{3\arrayrulewidth}
  \end{tabularx}
\end{table}

\subsubsection{Pruning Different modules} \label{sec:details_different_modules}

In Section \textcolor{red}{3.4}, we note that pruning different modules exerts heterogeneous effects on the overall performance of Mamba-based LLMs, with sensitivity varying markedly across modules.
Specifically, pruning the \texttt{in\_proj} module precipitates a precipitous decline in model performance, and pruning the \texttt{out\_proj} module similarly induces significant degradation, whereas remaining modules demonstrate higher resilience to parameter removal. 

Within the Mamba architecture, the \texttt{in\_proj} and \texttt{out\_proj} modules serve as the principal input projection and output transformation layers, respectively, endowing them with high coupling and low redundancy that limit their prunability.
Conversely, other modules are characterized by extensive overparameterization, enabling redundant representations of analogous functionalities and yielding comparatively low Hessian curvature across their parameters. 

\begin{table}[!htbp]
\centering
\caption{ Performance analysis results for pruning different modules. In each row, the Module column denotes the component being pruned, with 50\% sparsity applied to the Mamba-370M model.
}
\vspace{1mm}
\resizebox{0.95\columnwidth}{!}{%
\renewcommand{\arraystretch}{1.3}
\begin{tabular}{c|*{9}{W{c}{3em}}}
\Xhline{3\arrayrulewidth}
Module  & Wiki. $\downarrow$ & PTB $\downarrow$ & C4 $\downarrow$ 
      & OBQA $\uparrow$ & PIQA $\uparrow$ & ARC-e $\uparrow$ & ARC-c $\uparrow$
      & WinoG $\uparrow$ & Avg. $\uparrow$ \\ 
\Xhline{3\arrayrulewidth}
\texttt{conv1d}             & 14.46 & 23.78 & 19.52 & 30.80 & 68.61 & 55.13 & 27.30 & 55.01 & 47.37 \\ 
\texttt{in\_proj}    & 16.28 & 27.23 & 22.68 
                   & 30.40 & 66.43 & 51.56 & 26.88 & 55.25 & 46.10
 \\[0.2em]
\texttt{x\_proj}            & 14.35 & 23.55 & 19.39
 & 30.40 & 68.55 & 54.59 & 27.90 & 55.64 & 47.42 \\ 
\texttt{dt\_proj}     & 14.49 & 23.88 & 19.56
                   & 30.80 & 68.39 & 54.50 & 28.75 & 54.78 & 47.44
\\[0.2em]

\texttt{out\_proj}           & 15.19 & 25.45 & 21.47 & 31.00 & 66.87 & 54.08 & 27.56 & 56.12 & 47.13

 \\
\Xhline{3\arrayrulewidth}
\end{tabular}%
}
\label{tab:timestep}
\end{table}

\subsubsection{Results of Pruning SSM Module at High Sparsity} \label{sec:details_more_ssm}

We further compare our method against magnitude pruning (MP), Mamba-Shedder, and SparseGPT across a range of sparsity levels. 
The pruning results for these methods are reported on Mamba-130M, Mamba-370M, Mamba-790M, and Mamba-1.4B. 
We evaluate the perplexity of each pruned model on WikiText-2, PTB, and C4, and measure task accuracy on OpenBookQA, PIQA, ARC-Easy, ARC-Challenge, and Winogrande. 
As summarized in Table~\ref{tab:ssm-0.4}, \ref{tab:ssm-0.6}, \ref{tab:ssm-0.7}, \ref{tab:ssm-0.8}, our approach consistently outperforms all baselines at every sparsity level, thereby demonstrating its robustness.

\begin{table}[t]
\centering
\caption{Performance analysis for one-shot unstructured pruning of SSM modules in Mamba models (130M $\sim$ 1.4B) at $40\%$ sparsity. 
Here, $\downarrow$ lower metrics reflect better outcomes, and $\uparrow$ denotes higher metrics reflect better outcomes.}
\vspace{1mm}
\resizebox{0.95\columnwidth}{!}{%
\renewcommand{\arraystretch}{1.3}
\begin{tabular}{c|c|*{9}{W{c}{3em}}}
\Xhline{3\arrayrulewidth}
Model & Method  & Wiki. $\downarrow$ & PTB $\downarrow$ & C4 $\downarrow$ 
      & OBQA $\uparrow$ & PIQA $\uparrow$ & ARC-e $\uparrow$ & ARC-c $\uparrow$
      & WinoG $\uparrow$ & Avg. $\uparrow$ \\ \hline
\multirow{5}{*}{Mamba-130M} 
  & Dense          & 20.60 & 32.75 & 25.66 & 28.60 & 63.28 & 48.02 & 24.40 & 52.5 & 43.36 \\ 
  & MP \citep{magnitude}            & 218.7 & 304.86 & 107.77 & 28.20 & \underline{60.72} & 40.57 & 23.29 & \underline{51.85} & \underline{40.93} \\ 
  & Mamba-Shedder \citep{mamba-shedder}  & 275.3 & 506.6 & 222.8 & 25.00 & 55.11 & 34.89 & 22.10 & 49.72 & 37.37 \\
  & SparseGPT \citep{sparsegpt}     & \underline{165.0} & \underline{211.3} & \underline{87.22} & \underline{28.80} & 59.96 & \underline{40.66} & \textbf{24.74} & 50.43 & 40.92 \\
  & \modelname     & \textbf{25.23} & \textbf{42.79} & \textbf{29.45} 
                   & \textbf{30.00} & \textbf{62.57} & \textbf{46.00} & \underline{24.23}
                   & \textbf{52.49} & \textbf{43.06} \\ \hline 
\multirow{5}{*}{Mamba-370M}
  & Dense          & 14.32 & 23.46 & 19.37 & 31.00 & 68.34 & 54.97 & 27.90 & 55.25 & 47.49 \\ 
  & MP \citep{magnitude}             & \underline{149.8} & \underline{264.8} & \underline{70.17} & 31.00 & \underline{65.89} & \underline{51.22} & 25.77 & 51.85 & 45.15 \\ 
  & Mamba-Shedder \citep{mamba-shedder}   & 195.5 & 310.6 & 137.9 & 26.20 & 56.80 & 30.60 & 22.10 & 49.64 & 37.07 \\
  & SparseGPT \citep{sparsegpt}     & 2.8e4 & 4.6e6 & 6367 & \textbf{31.80} & \underline{65.89} & 50.84 & \underline{26.54} & \underline{53.04} & \underline{45.62} \\
  & \modelname     & \textbf{16.90} & \textbf{27.72} & \textbf{22.28} 
                   & \underline{31.60} & \textbf{68.61} & \textbf{53.91} & \textbf{27.22} & \textbf{55.64} & \textbf{47.40} \\ \hline 
\multirow{5}{*}{Mamba-790M} 
  & Dense          & 11.96 & 18.45 & 16.62 & 33.80 & 72.63 & 61.07 & 29.44 & 56.27 & 50.64 \\ 
  & MP \citep{magnitude}            & 97.37 & 150.4 & 53.35 & 32.40 & 68.66 & 54.17 & \underline{27.65} & 55.33 & 47.64 \\ 
  & Mamba-Shedder \citep{mamba-shedder}   & 75.51 & 109.5 & 78.93 & \textbf{33.60} & \underline{71.06} & \underline{56.57} & 27.39 & \underline{55.72} & \underline{48.87} \\
  & SparseGPT \citep{sparsegpt}     & \underline{36.14} & \underline{81.62} & \underline{34.13} & \underline{32.80} & 68.34 & 54.42 & 27.47 & 54.93 & 47.59 \\
  & \modelname     & \textbf{13.81} & \textbf{22.47} & \textbf{18.62} 
                   & 32.60 & \textbf{72.85} & \textbf{58.96} & \textbf{27.90} & \textbf{57.14} & \textbf{49.89} \\ \hline 
\multirow{5}{*}{Mamba-1.4B} 
  & Dense          & 10.75 & 17.05 & 15.17 & 36.40 & 73.88 & 65.57 & 32.85 & 61.17 & 53.98 \\
  & MP \citep{magnitude}             & 49.99 & 84.70 & 34.14 & 34.60 & 70.35 & 59.68 & 27.82 & 56.04 & 49.70 \\ 
  & Mamba-Shedder \citep{mamba-shedder}  & 120.6 & 179.5 & 109.7 & 26.40 & 60.45 & 39.86 & 22.95 & 52.41 & 40.41 \\
  & SparseGPT \citep{sparsegpt}     & \underline{32.39} & \underline{49.87} & \underline{28.86} & \textbf{36.20} & \underline{72.36} & \underline{61.49} & \underline{31.48} & \underline{57.30} & \underline{51.77} \\
  & \modelname     & \textbf{13.03} & \textbf{34.53} & \textbf{17.15} 
                   & \underline{35.20} & \textbf{73.56} & \textbf{64.14} & \textbf{32.59} & \textbf{58.72} & \textbf{52.84} \\
\Xhline{3\arrayrulewidth}
\end{tabular}%
}
\label{tab:ssm-0.4}
\end{table}

\begin{table}[t]
\centering
\caption{Performance analysis for one-shot unstructured pruning of SSM modules in Mamba models (130M $\sim$ 1.4B) at $60\%$ sparsity. 
Here, $\downarrow$ lower metrics reflect better outcomes, and $\uparrow$ denotes higher metrics reflect better outcomes.}
\vspace{1mm}
\resizebox{0.95\columnwidth}{!}{%
\renewcommand{\arraystretch}{1.3}
\begin{tabular}{c|c|*{9}{W{c}{3em}}}
\Xhline{3\arrayrulewidth}
Model & Method  & Wiki. $\downarrow$ & PTB $\downarrow$ & C4 $\downarrow$ 
      & OBQA $\uparrow$ & PIQA $\uparrow$ & ARC-e $\uparrow$ & ARC-c $\uparrow$
      & WinoG $\uparrow$ & Avg. $\uparrow$ \\ \hline
\multirow{5}{*}{Mamba-130M} 
  & Dense          & 20.60 & 32.75 & 25.66 & 28.60 & 63.28 & 48.02 & 24.40 & 52.50 & 43.36 \\ 
  & MP \citep{magnitude}            & \underline{1034} & \underline{1605} & \underline{351.7} & 26.00 & \underline{55.55} & \underline{33.42} & 22.10 & 49.96 & \underline{37.41} \\ 
  & Mamba-Shedder \citep{mamba-shedder}  & 3219 & 4998 & 1503 & 25.80 & 54.46 & 29.00 & \underline{23.72} & \underline{50.04} & 36.60 \\
  & SparseGPT \citep{sparsegpt}     & 5.0e4 & 1.4e4 & 2.4e4 & \underline{26.20} & 52.45 & 26.85 & 23.55 & 49.80 & 35.77 \\
  & \modelname     & \textbf{33.74} & \textbf{59.47} & \textbf{35.02} & \textbf{31.20} & \textbf{62.35} & \textbf{45.16} & \textbf{24.06} & \textbf{50.67} & \textbf{42.69} \\ \hline
\multirow{5}{*}{Mamba-370M}
  & Dense          & 14.32 & 23.46 & 19.37 & 31.00 & 68.34 & 54.97 & 27.90 & 55.25 & 47.49 \\ 
  & MP \citep{magnitude}             & 386.2 & 747.6 & \underline{141.6} & 26.40 & 58.05 & 38.64 & 21.59 & 49.64 & 38.86 \\ 
  & Mamba-Shedder \citep{mamba-shedder}   & 463.3 & \underline{561.6} & 307.0 & 25.00 & 54.03 & 28.91 & 23.63 & 49.72 & 36.26 \\
  & SparseGPT \citep{sparsegpt}     & \underline{360.2} & 1455 & 324.7 & \underline{30.00} & \underline{58.87} & \underline{40.07} & \underline{23.89} & \textbf{53.28} & \underline{41.22} \\
  & \modelname     & \textbf{22.65} & \textbf{38.37} & \textbf{28.10} & \textbf{31.80} & \textbf{66.76} & \textbf{49.45} & \textbf{27.82} & \underline{52.80} & \textbf{45.73} \\ \hline
\multirow{5}{*}{Mamba-790M} 
  & Dense          & 11.96 & 18.45 & 16.62 & 33.80 & 72.63 & 61.07 & 29.44 & 56.27 & 50.64 \\ 
  & MP \citep{magnitude}            & \underline{255.6} & 502.5 & \underline{108.4} & 28.40 & 60.61 & 41.92 & 23.29 & 51.85 & 41.22 \\ 
  & Mamba-Shedder \citep{mamba-shedder}   & 353.5 & \underline{358.3} & 283.5 & 26.60 & 54.95 & 32.58 & 23.04 & 49.96 & 37.43 \\
  & SparseGPT \citep{sparsegpt}     & 1033 & 3630 & 897.5 & \underline{31.40} & \underline{65.40} & \underline{51.26} & \underline{24.74} & \underline{53.67} & \underline{45.29} \\
  & \modelname     & \textbf{18.45} & \textbf{30.29} & \textbf{22.64} & \textbf{31.60} & \textbf{69.31} & \textbf{56.65} & \textbf{26.37} & \textbf{55.80} & \textbf{47.95} \\ \hline
\multirow{5}{*}{Mamba-1.4B} 
  & Dense          & 10.75 & 17.05 & 15.17 & 36.40 & 73.88 & 65.57 & 32.85 & 61.17 & 53.98 \\
  & MP \citep{magnitude}             & 150.9 & 322.3 & \underline{67.64} & 30.20 & 62.73 & 47.47 & 25.43 & 50.99 & 43.36 \\ 
  & Mamba-Shedder \citep{mamba-shedder}  & 370.4 & 481.4 & 281.4 & 26.80 & 55.55 & 33.67 & 23.29 & 50.67 & 38.00 \\
  & SparseGPT \citep{sparsegpt}     & \underline{110.3} & \underline{209.2} & 70.36 & \textbf{34.60} & \textbf{69.91} & \textbf{58.59} & \underline{27.99} & \underline{53.75} & \textbf{48.97} \\
  & \modelname     & \textbf{26.52} & \textbf{53.15} & \textbf{22.82} & \underline{32.00} & \underline{69.26} & \underline{56.90} & \textbf{28.16} & \textbf{56.12} & \underline{48.49} \\
\Xhline{3\arrayrulewidth}
\end{tabular}%
}
\label{tab:ssm-0.6}
\end{table}

\begin{table}[t]
\centering
\caption{Performance analysis for one-shot unstructured pruning of SSM modules in Mamba models (130M $\sim$ 1.4B) at $70\%$ sparsity.  
Here, $\downarrow$ lower metrics reflect better outcomes, and $\uparrow$ denotes higher metrics reflect better outcomes.}
\vspace{1mm}
\resizebox{0.95\columnwidth}{!}{%
\renewcommand{\arraystretch}{1.3}
\begin{tabular}{c|c|*{9}{W{c}{3em}}}
\Xhline{3\arrayrulewidth}
Model & Method  & Wiki. $\downarrow$ & PTB $\downarrow$ & C4 $\downarrow$ 
      & OBQA $\uparrow$ & PIQA $\uparrow$ & ARC-e $\uparrow$ & ARC-c $\uparrow$
      & WinoG $\uparrow$ & Avg. $\uparrow$ \\ \hline
\multirow{5}{*}{Mamba-130M} 
  & Dense          & 20.60 & 32.75 & 25.66 & 28.60 & 63.28 & 48.02 & 24.40 & 52.50 & 43.36 \\ 
  & MP \citep{magnitude}            & \underline{1248} & \underline{1802} & \underline{407.0} & 24.80 & \underline{54.13} & \underline{30.68} & 24.32 & \textbf{52.49} & \underline{37.28} \\ 
  & Mamba-Shedder \citep{mamba-shedder}  & 5845 & 1.2e4 & 3775 & \underline{26.80} & 51.85 & 26.56 & \textbf{24.57} & \underline{50.67} & 36.09 \\
  & SparseGPT \citep{sparsegpt}     & 1.1e5 & 6.7e4 & 1.8e5 & 24.20 & 51.47 & 25.59 & \underline{24.40} & 50.36 & 35.20 \\
  & \modelname     & \textbf{43.72} & \textbf{72.05} & \textbf{40.82} &
                     \textbf{30.00} & \textbf{60.99} & \textbf{41.41} & 22.87 &
                     50.59 & \textbf{41.17} \\ \hline
\multirow{5}{*}{Mamba-370M} 
  & Dense          & 14.32 & 23.46 & 19.37 & 31.00 & 68.34 & 54.97 & 27.90 & 55.25 & 47.49 \\ 
  & MP \citep{magnitude}            & \underline{497.3} & \underline{925.2} & \underline{174.4} & 25.60 & 56.91 & 36.70 & 20.05 & \underline{51.30} & 38.11 \\ 
  & Mamba-Shedder \citep{mamba-shedder}  & 1029 & \underline{933.0} & 625.7 & 26.80 & 52.67 & 27.95 & \underline{23.63} & 50.28 & 36.26 \\
  & SparseGPT \citep{sparsegpt}     & 7.8e4 & 5.5e4 & 7.3e4 & \textbf{27.80} & \underline{59.30} & \underline{39.23} & 22.61 & 50.36 & \underline{39.86} \\
  & \modelname     & \textbf{28.28} & \textbf{44.93} & \textbf{33.28} &
                     \underline{27.00} & \textbf{65.56} & \textbf{49.03} & \textbf{24.06} &
                     \textbf{53.35} & \textbf{43.80} \\ \hline
\multirow{5}{*}{Mamba-790M} 
  & Dense          & 11.96 & 18.45 & 16.62 & 33.80 & 72.63 & 61.07 & 29.44 & 56.27 & 50.64 \\ 
  & MP \citep{magnitude}            & \underline{341.2} & 655.9 & \underline{139.6} & 27.00 & 57.83 & 38.85 & \textbf{24.57} & \underline{51.22} & 39.90 \\ 
  & Mamba-Shedder \citep{mamba-shedder}  & 353.5 & \underline{358.3} & 283.5 & 26.60 & 54.95 & 32.58 & 23.04 & 49.96 & 37.43 \\
  & SparseGPT \citep{sparsegpt}     & 1.9e5 & 2.4e7 & 2.7e5 & \underline{27.60} & \underline{61.32} & \underline{39.10} & \underline{24.49} & \textbf{52.96} & \underline{41.09} \\
  & \modelname     & \textbf{21.62} & \textbf{43.00} & \textbf{25.59} &
                     \textbf{32.00} & \textbf{67.14} & \textbf{51.01} & 23.98 &
                     50.83 & \textbf{44.99} \\ \hline
\multirow{5}{*}{Mamba-1.4B} 
  & Dense          & 10.75 & 17.05 & 15.17 & 36.40 & 73.88 & 65.57 & 32.85 & 61.17 & 53.98 \\
  & MP \citep{magnitude}            & \underline{180.8} & \underline{378.5} & \underline{80.16} & 28.80 & 59.58 & 41.67 & 23.55 & \underline{51.07} & 40.93 \\ 
  & Mamba-Shedder \citep{mamba-shedder}  & 805.1 & 796.6 & 541.7 & 25.40 & 54.08 & 29.50 & 24.06 & 49.09 & 36.43 \\
  & SparseGPT \citep{sparsegpt}     & 452.5 & 602.9 & 253.7 & \underline{31.20} & \underline{66.27} & \textbf{54.00} & \underline{24.40} & 50.36 & \underline{45.24} \\
  & \modelname     & \textbf{42.46} & \textbf{74.21} & \textbf{30.24} &
                     \textbf{31.40} & \textbf{66.92} & \underline{51.30} & \textbf{27.39} &
                     \textbf{53.43} & \textbf{46.09} \\
\Xhline{3\arrayrulewidth}
\end{tabular}%
}
\label{tab:ssm-0.7}
\end{table}

\begin{table}[t]
\centering
\caption{Performance analysis for one-shot unstructured pruning of SSM modules in Mamba models (130M $\sim$ 1.4B) at $80\%$ sparsity.  
Here, $\downarrow$ lower metrics reflect better outcomes, and $\uparrow$ denotes higher metrics reflect better outcomes.}
\vspace{1mm}
\resizebox{0.95\columnwidth}{!}{%
\renewcommand{\arraystretch}{1.3}
\begin{tabular}{c|c|*{9}{W{c}{3em}}}
\Xhline{3\arrayrulewidth}
Model & Method  & Wiki. $\downarrow$ & PTB $\downarrow$ & C4 $\downarrow$ 
      & OBQA $\uparrow$ & PIQA $\uparrow$ & ARC-e $\uparrow$ & ARC-c $\uparrow$
      & WinoG $\uparrow$ & Avg. $\uparrow$ \\ \hline
\multirow{5}{*}{Mamba-130M} 
  & Dense          & 20.60 & 32.75 & 25.66 & 28.60 & 63.28 & 48.02 & 24.40 & 52.50 & 43.36 \\ 
  & MP \citep{magnitude}            & \underline{1297} & \underline{1870} & \underline{420.4} & 24.00 & 52.18 & \underline{31.31} & \textbf{24.32} & 50.51 & 36.46 \\ 
  & Mamba-Shedder \citep{mamba-shedder}  & 2.6e4 & 5.9e4 & 2.2e4 & \underline{26.20} & 51.69 & 28.03 & \underline{23.89} & \textbf{52.01} & 36.36 \\
  & SparseGPT \citep{sparsegpt}     & 2.6e21 & 5.7e22 & 2.7e23 & 24.80 & \underline{55.98} & 30.60 & 23.38 & \underline{51.30} & \underline{37.21} \\
  & \modelname     & \textbf{65.90} & \textbf{124.1} & \textbf{57.45} 
                   & \textbf{28.80} & \textbf{57.02} & \textbf{38.43} & 23.21
                   & 49.96 & \textbf{39.48} \\ \hline
\multirow{5}{*}{Mamba-370M}
  & Dense          & 14.32 & 23.46 & 19.37 & 31.00 & 68.34 & 54.97 & 27.90 & 55.25 & 47.49 \\ 
  & MP \citep{magnitude}             & \underline{538.2} & 983.0 & \underline{191.0} & 25.20 & 53.16 & 31.99 & 22.61 & 49.49 & 36.49 \\ 
  & Mamba-Shedder \citep{mamba-shedder}   & 3191 & \underline{933.0} & 1848 & \underline{27.80} & 52.34 & 26.52 & \textbf{24.06} & 51.14 & 36.37 \\
  & SparseGPT \citep{sparsegpt}     & 1.1e5 & 1.2e5 & 1.0e5 & 27.40 & \underline{56.26} & \underline{34.93} & \underline{23.38} & \textbf{53.83} & \underline{39.16} \\
  & \modelname     & \textbf{51.58} & \textbf{90.87} & \textbf{50.36} 
                   & \textbf{30.20} & \textbf{58.43} & \textbf{42.21} & 23.29 & \underline{51.54} & \textbf{41.14} \\ \hline
\multirow{5}{*}{Mamba-790M} 
  & Dense          & 11.96 & 18.45 & 16.62 & 33.80 & 72.63 & 61.07 & 29.44 & 56.27 & 50.64 \\ 
  & MP \citep{magnitude}            & \underline{402.7} & \underline{738.5} & \underline{160.1} & 25.80 & \underline{56.80} & 36.66 & 22.70 & 49.41 & 38.27 \\ 
  & Mamba-Shedder \citep{mamba-shedder}   & 1891 & 2121 & 1277 & 25.40 & 51.69 & 28.28 & \underline{24.40} & 48.15 & 35.58 \\
  & SparseGPT \citep{sparsegpt}     & 1.7e8 & 4.3e8 & 2.1e8 & \underline{27.40} & \underline{56.80} & \underline{36.70} & 23.12 & \textbf{50.91} & \underline{38.99} \\
  & \modelname     & \textbf{33.75} & \textbf{67.97} & \textbf{34.85} 
                   & \textbf{31.40} & \textbf{63.77} & \textbf{47.43} & \textbf{24.49} & \underline{50.28} & \textbf{43.47} \\ \hline
\multirow{5}{*}{Mamba-1.4B} 
  & Dense          & 10.75 & 17.05 & 15.17 & 36.40 & 73.88 & 65.57 & 32.85 & 61.17 & 53.98 \\
  & MP \citep{magnitude}             & \underline{227.4} & \underline{438.6} & \underline{101.4} & 25.00 & 56.04 & 34.81 & 22.61 & \textbf{53.12} & 38.31 \\ 
  & Mamba-Shedder \citep{mamba-shedder}  & 2260 & 2236 & 1405 & 26.80 & 51.20 & 28.87 & \textbf{27.13} & 51.14 & 37.03 \\
  & SparseGPT \citep{sparsegpt}     & 5.7e11 & 2.6e13 & 3.1e14 & \underline{28.20} & \underline{59.36} & \underline{43.22} & 23.38 & 48.93 & \underline{40.62} \\
  & \modelname     & \textbf{88.93} & \textbf{144.8} & \textbf{45.34} 
                   & \textbf{30.60} & \textbf{62.35} & \textbf{45.08} & \underline{24.57} & \underline{51.38} & \textbf{42.80} \\
\Xhline{3\arrayrulewidth}
\end{tabular}%
}
\label{tab:ssm-0.8}
\end{table}

\section{Further Discuss} \label{sec:discuss}

\noindent\textbf{Limitations.} 
Our proposed method represents the first work to extend the OBS framework to Mamba-based LLMs.
While it can be naturally extended to structured pruning of the SSM module, further work is required to develop a one‐shot, second‐order information–based structured pruning strategy that effectively accelerates the entire model. 
In our preliminary structured‐pruning extension, we achieved a 1.72× speed‐up on the SSM module, yet the end-to-end inference acceleration of the full model remains modest. 
Moreover, since our experiments were conducted on open-source Mamba model series, their deployment may inherently entail ethical and safety risks.

\noindent\textbf{Broader Impact.} 
Our proposed method effectively reduces parameter redundancy in Mamba-based LLMs, yielding a leaner network representation that requires fewer floating-point operations during inference.
As a result, these pruned models can be deployed with lower computational cost, both in terms of GPU hours and energy consumption, thereby democratizing access to state-of-the-art LLM capabilities for academic, industrial, and edge computing environments. 
Moreover, by curtailing the extensive resource demands traditionally associated with LLM inference, our approach contributes to a reduction in the cumulative electricity usage and associated carbon emissions of LLM workloads. 
In doing so, it supports the broader agenda of sustainable AI by mitigating the environmental and climate impacts of deploying LLMs at scale.

\end{document}